\documentclass[runningheads]{llncs}
\usepackage{graphicx}
\usepackage{svg}
\usepackage{microtype}
\usepackage{tikz}
\usepackage{comment}
\usepackage{amsmath,amssymb} 
\usepackage{color}
\usepackage[arrowdel]{physics}
\usepackage{booktabs}
\usepackage{multirow}
\usepackage{pgfplots, pgfplotstable}
\pgfplotsset{compat=1.9}
\usepackage[pagebackref,breaklinks,colorlinks]{hyperref}
\usepackage[accsupp]{axessibility}  
\usepackage{floatrow}
\newfloatcommand{capbtabbox}{table}[][\FBwidth]
\usepackage{float}
\floatstyle{plaintop}
\restylefloat{table}
\usepackage{blindtext}

\begin{document}
\pagestyle{headings}
\mainmatter
\def\ECCVSubNumber{1038}

\title{Revisiting Consistency Regularization for Semi-supervised Change Detection in Remote Sensing Images}

\titlerunning{ Consistency Regularization for
Semi-supervised Change Detection}
%
\author{Wele Gedara Chaminda Bandara \and
Vishal M. Patel}
\authorrunning{W.G.C. Bandara et al.}
%
\institute{Johns Hopkins University, USA\\
\email{\{wbandar1, vpatel36\}@jhu.edu}}

\maketitle

\begin{abstract}
Remote-sensing (RS) Change Detection (CD) aims to detect ``changes of interest'' from co-registered bi-temporal images. The performance of existing deep supervised CD methods is attributed to the large amounts of annotated data used to train the networks. However, annotating large amounts of remote sensing images is labor intensive and expensive, particularly with bi-temporal images, as it requires pixel-wise comparisons by a human expert. On the other hand, we often have access to unlimited unlabeled multi-temporal RS imagery thanks to ever-increasing earth observation programs. In this paper, we propose a simple yet effective way to leverage the information from unlabeled bi-temporal images to improve the performance of CD approaches. More specifically, we propose a semi-supervised CD model in which we formulate an unsupervised CD loss in addition to the supervised Cross-Entropy (CE) loss by constraining the output change probability map of a given unlabeled bi-temporal image pair to be consistent under the small random perturbations applied on the deep feature difference map that is obtained by subtracting their latent feature representations. Experiments conducted on two publicly available CD datasets show that the proposed semi-supervised CD method can reach closer to the performance of supervised CD even with access to as little as 10\% of the annotated training data. Code available at \href{https://github.com/wgcban/SemiCD}{https://github.com/wgcban/SemiCD}.
\end{abstract}
\vspace{-5mm}
\begin{figure}[!htb]
    \centering
    \includegraphics[width=\linewidth]{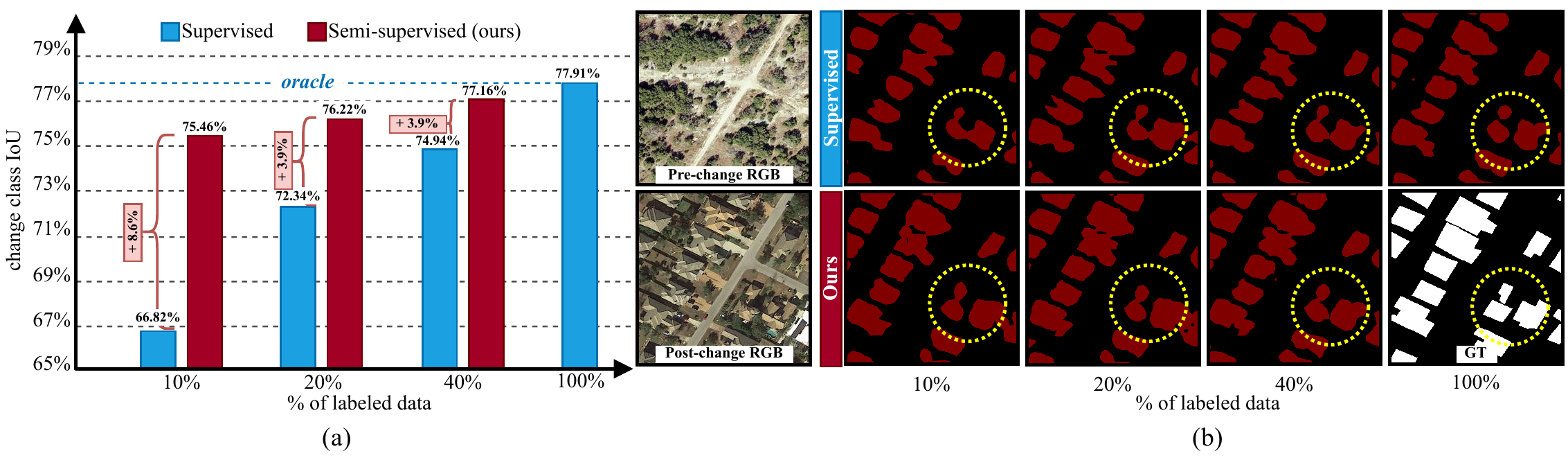}
    \caption{The efficiency of the proposed semi-supervised CD approach. By utilizing only 10\% of the \textit{training-set} of LEVIR~\cite{levid-cd} in a supervised manner and remaining 90\% as unsupervised data, our semi-supervised CD method is only 3.1\% below the result of the supervised CD method that utilizes 100\% of the LEVIR \textit{training-set}.}
    \label{fig:intro}
\end{figure}

\section{Introduction}
Change Detection (CD) is an important problem in remote sensing in which our goal is to identify \textit{relevant changes} in bi-temporal remote sensing images that can help in various applications such as damage assessment, urban expansion monitoring, resource management, and military surveillance~\cite{caye2019guided,changeformer,daudt2018fully}. Depending on the application, the changes we look for in bi-temporal images also vary. For example, in the context of military surveillance and disaster management, changes in buildings may indicate a developing threat or areas to focus on for disaster relief, respectively. Therefore, an ideal CD model should be able to detect all of these application-specific \textit{changes of interest} while avoiding  complex \textit{irrelevant changes} typically seen in remote sensing images such as changes caused by cloud cover, building shadows, different sensing angles, etc.~\cite{cd_survey}.

Various CD methods have been proposed to identify these \textit{relevant changes} efficiently and accurately. Of these methods, most are based on \textit{supervised learning} in which a model is trained in a supervised manner using a supervised loss function such as Cross-Entropy (CE) by utilizing annotated training samples. Even though supervised CD models can avoid complex \textit{irrelevant changes} to a great extent, they usually require an extensive collection of annotated training examples, which is expensive and often time-consuming - especially with bi-temporal remote-sensing images as it requires comparison of two images semantically as part of the annotation process~\cite{shen2021s2looking}. Furthermore, these supervised CD models usually over-fit to the training dataset resulting in a drastic drop in performance under the domain shift~\cite{wang2021loveda} - which is very common in remote sensing images due to the fact that different imaging sensors, different terrain conditions, seasonal variations, and different illumination conditions are possible when capturing images. On the contrary, unsupervised CD models either utilize unsupervised loss functions such as contrastive loss~\cite{contrastive_cd} and similarity-dissimilarity loss~\cite{bruzzone2000automatic,semi_metric_learning,celik2009unsupervised,melgani2002unsupervised,semi_SVM} to identify the change regions or generative models (like GANs or any other synthetic change generation pipeline)~\cite{unsup_gan} to synthetically create a training dataset with known change labels to train a CD model in a supervised manner. However, the main drawback of the aforementioned unsupervised CD methods is that they are very weak in distinguishing \textit{changes of interest} from \textit{irrelevant changes} as no such information is given to the CD model during training~\cite{bovolo2006theoretical}. 

To overcome the drawbacks of supervised and unsupervised CD methods mentioned above, we propose a semi-supervised CD framework. We begin our study by examining the \textit{cluster assumption} for consistency-based semi-supervised learning~\cite{chapelle2009semi}, which states that the boundary between change and no-change classes should be aligned with the boundary of low-density regions. If such cluster assumption remains for CD, we can utilize consistency training to leverage the information from unlabeled bi-temporal remote sensing images by enforcing the model’s predictions to be invariant over small random perturbations applied to the inputs. We demonstrate that, for CD, such cluster assumption does not hold in multi-temporal image-domain, but rather it holds in the deep feature difference-domain. Hence, we subsequently apply small random perturbations to the deep feature difference map and make the resulting change map from the deep network consistent with the one without any perturbation. Combining this unsupervised consistency loss with the supervised CE loss, we can effectively leverage the information from labeled and unlabeled multi-temporal remote sensing images to improve the CD performance compared to training the network only in a supervised way, as shown in Fig. \ref{fig:intro}. We can see that the proposed semi-supervised method can almost reach the Intersection over Union (IoU) of the supervised method by only utilizing 10\% of the training data that the supervised model uses for training and the remaining 90\% of data in an unsupervised way via the proposed consistency regularization approach. Furthermore, since the proposed semi-supervised model is robust under the small perturbations on the feature-difference map, it can also perform relatively better on out-of-distribution (i.e., different dataset) samples as we demonstrate later in this paper. 

In summary, we make the following contributions:
\begin{itemize}
    \item We propose a semi-supervised CD paradigm based on consistency regularization. The proposed approach can effectively leverage the information from freely-available, unlabeled, multi-temporal, remote-sensing images to enhance the CD performance. To the best of our knowledge, this is the first work which proposes a deep semi-supervised approach for CD.
    
    \item We reformulate the fundamental assumption of consistency-based regularization - the \textit{cluster assumption} for CD. We empirically show that the cluster assumption does not hold in the bi-temporal image domain but rather it holds in the deep feature difference-domain.

    \item Our method of leveraging information from the unlabeled, multi-temporal remote-sensing images is simple. Since, the cluster assumption holds in the deep feature difference domain, we apply small random perturbations to the deep feature difference map of a given bi-temporal image pair and make the deep network's prediction consistent with the one without any perturbation. Furthermore, this approach makes our CD model produce  better change maps even under small perturbations that appear in the feature difference maps when testing on the out-of-distribution data.
    
    \item Finally, we conduct extensive experiments including cross-dataset to demonstrate the effectiveness of our method.
\end{itemize}

\vspace{-5mm}
\section{Related works}
\label{sec:related_work}
\subsubsection{Supervised CD.} Given a set of bi-temporal image pairs and their corresponding ground-truth change labels, supervised CD methods train a model by minimizing the Cross-Entropy (CE) loss over the training dataset~\cite{changeformer,bit}. These supervised CD networks are mainly based on convolutional neural networks~\cite{daudt2018fully,daudt2018urban,zhang2021escnet,xu2020pseudo,qu2021change,chen2019change,varghese2018changenet,toker2022dynamicearthnet}, attention/transformer architectures~\cite{changeformer}, or the combination of both~\cite{bit,chen2020dasnet,fang2021snunet,zhang2020deeply,diakogiannis2021looking}. Since it is required to perform some form of comparison between the pre-change and post-change image to obtain the change map, the existing methods either adopt: (1) \textit{Early fusion}~\cite{wiratama2019fusion,zhang2020coarse,song2021efficient,daudt2018urban,daudt2018fully}, which processes fused pre-change and post-change images along the spectral dimension through the CD network, (2) \textit{Siamese-concatenation}~\cite{zhang2021hdfnet,daudt2018fully}, in which hidden features of pre-change and post-change images from a shared encoder are concatenated and then  processed through the decoder to obtain the change map, or (3) \textit{Siamese-difference}~\cite{changeformer,bit,daudt2018fully}, in which the absolute difference of the hidden features of a pre-change and a post-change are processed through the decoder to obtain the change map. Among these three methods, the Siamese difference has been shown to produce better results in many cases, making it the basic architecture for the CD problem. However, due to the limited annotated training-data, most of these architectures rely on \textit{transfer learning}~\cite{bit,8937755} as the starting point of the CD network (or part of the network like encoder) that have been trained on a larger dataset for a different problem like ImageNet classification~\cite{krizhevsky2012imagenet}. However, this approach is limited in the sense that it assumes similarities between the natural images and remote sensing data \cite{manas2021seasonal}, and is unable to utilize freely-available, unlabeled remote sensing data particularly in to the CD problem. In order to fully utilize the available unlabeled remote sensing data into the CD problem and to reduce the dependency on a large number of annotated training samples, there is a growing interest among researchers to explore semi-supervised methods for CD.
\vspace{-6mm}
\subsubsection{Semi-supervised CD methods on individual image pairs.} Semi-supervised learning introduces regularization on unlabeled data to improve supervised learning, making it possible to train a CD model with limited annotated data without outfitting it on the labeled-training set. How to effectively harness the information from unlabeled bi-temporal images to improve the CD performance is a long-standing question, and researchers have attempted various methods to answer this question in the past. Among these, Bovolo \textit{et al.} \cite{semi_SVM}  proposed a semi-supervised CD on multispectral images using a properly defined S\textsuperscript{3}VM technique and an unsupervised model selection strategy based on a similarity measure.  Later, Chen \textit{et al.}~\cite{semi_GP} introduced the Gaussian process for semi-supervised CD in which difference images are first generated, and then both the labeled and unlabeled data was exploited by a probabilistic GP classifier. In order to overcome the shortcomings of the GP classifier and include the spatial contextual information, MRF regularization was employed by introducing edge information and high-order potentials. Furthermore, Ghosh \textit{et al.}~\cite{semi_SOFM} proposed a semi-supervised CD method based on a Self-Organizing Feature Map (SOFM), where only a few labeled patterns were utilized to initialize the SOFM network, and fuzzy set theory was then employed determine the membership values of the unlabeled data. Apart from these methods, Yuan \textit{et al.}~\cite{semi_metric_learning} introduced the metric learning to semi-supervised CD. In this method, first, a proper distance metric was learned so that the no-change class pixels are mapped closely to each other, while pixels from the change class are mapped further apart as much as possible. Next, the unlabeled data were incorporated into the CD problem via a Laplacian regularized framework. However, all of these semi-supervised CD methods mentioned above only investigate the \textit{individual image pairs} and cannot leverage the information from a \textit{large amount of unlabeled images}, which is essential for large-scale real-world applications.
\vspace{-5mm}
\subsubsection{Semi-supervised CD with a large amount of unlabeled images.} \label{sec:sota_semi} Only a very few works have focused on CD with a large amount of unlabeled RS images. In a more recent work,  Peng \textit{et al.}~\cite{SemiCDNet} proposed a solution to the above problem through a Generative Adversarial Network (GAN). The proposed method is based on two assumptions: the segmentation and entropy maps between the labeled and unlabeled data must follow the same distribution. Subsequently, two discriminators are adopted to reinforce these two assumptions. However, we argue that these two assumptions do not hold when the labeled and unlabeled change maps have different distributions and thus preventing us from utilizing images from another dataset to the CD problem. In contrast, our method enforces consistency between the network's predictions under small random perturbations applied in the hidden feature difference domain, allowing us to combine unlabeled RS images to the training process even from another domain, resulting in better generalization performance over multiple datasets.
\vspace{-5mm}
\subsubsection{Consistency regularization.} Among the many approaches available for semi-supervised learning, consistency training has shown state-of-the-art performance in different applications such as classification~\cite{oliver2018realistic,sohn2020fixmatch,verma2019interpolation}, semantic segmentation~\cite{ouali2020semi,mittal2019semi,lai2021semi}, object detection~\cite{jeong2019consistency}, and image-to-image translation~\cite{Wang_2020_CVPR,mustafa2020transformation}. The consistency regularization is inspired by two interconnected assumptions: the \textit{cluster assumption} and the \textit{low-density separation assumption}. The \textit{cluster assumption} states that if two samples belong to the same cluster in the input distribution, they are likely to belong to the same class, which indirectly infers the \textit{low-density assumption} which states that  the decision boundary separating the two classes should lie in the low-density regions~\cite{chapelle2009semi}. Thus, consistency regularization methods enforce the low-density separation assumption by encouraging the deep network $f$ to produce invariant prediction $f(u) = f(u+\delta)$ for random perturbation $\delta$ applied on unlabeled data point $u$. However, applying this formulation directly to the semi-supervised CD problem is challenging because adding perturbations to the input bi-temporal images could introduce new changes that may confuse the CD network during training and can result in poor CD performance. We suspect this is why consistency regularization has not been explored in the context of CD. In this paper, we show that the semi-supervised CD can also benefit from consistency regularization with impressive results by reformulating it appropriately to the CD problem.
\vspace{-4mm}
\section{Method}
\vspace{-3mm}
\subsection{The \textit{cluster assumption} in CD}
\label{sec:cluster_assumption}
\begin{figure}[!htb]
    \centering
    \includegraphics[width=\linewidth]{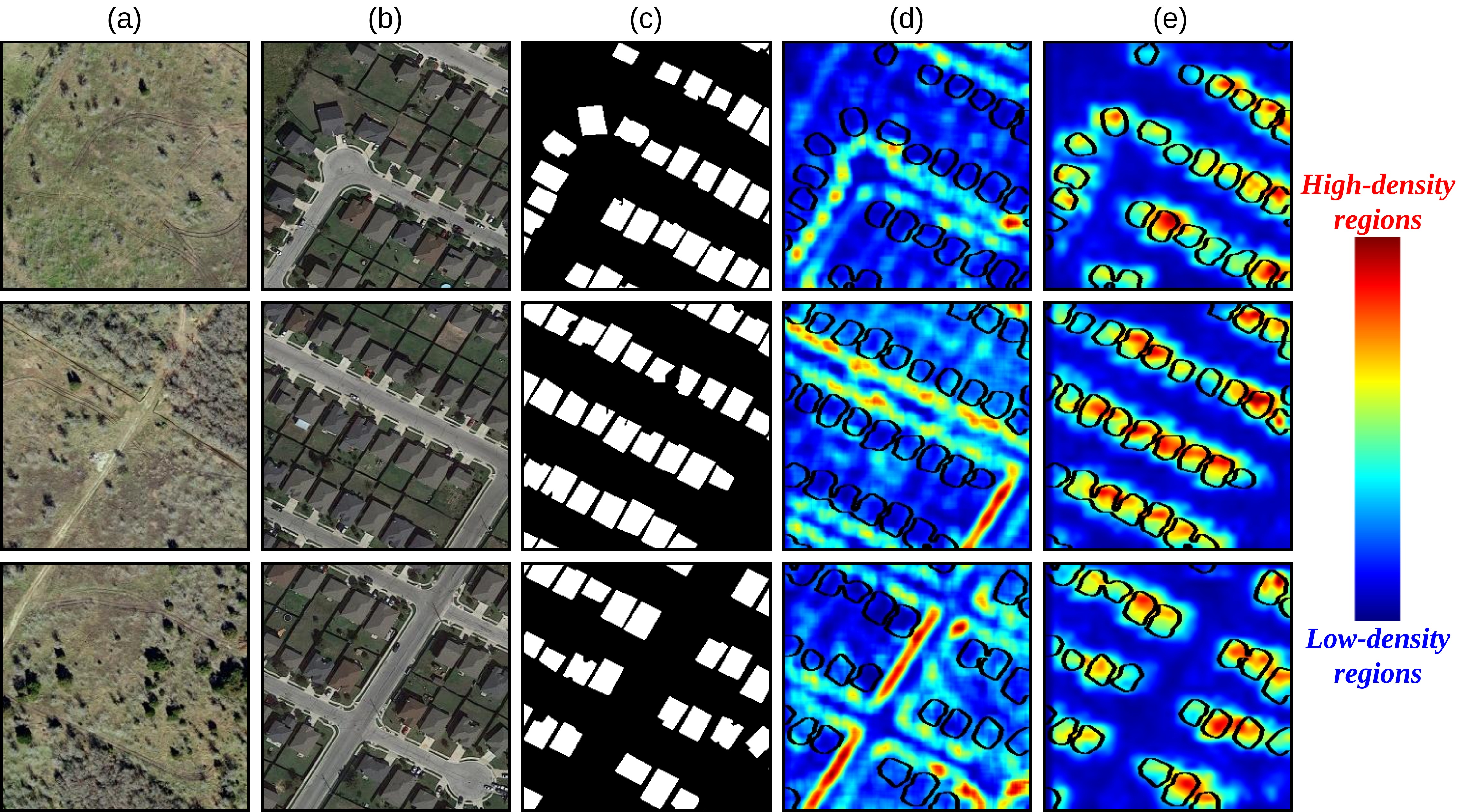}
    \caption{The cluster assumption in CD. \textbf{(a)} Pre-change images. \textbf{(b)} Post-change images. \textbf{(c)} Ground-truth change maps, where white-pixels corresponds to change regions and black-pixels corresponds to no-change regions. \textbf{(d)} The cluster assumption in the input bi-temporal image domain. \textbf{(e)} The cluster assumption in the hidden feature difference domain. Note that the regions closer to \textcolor{red}{red} color correspond to \textcolor{red}{high-density} regions while those with \textcolor{blue}{blue} color correspond to \textcolor{blue}{low-density} regions. The black lines indicate the segmentation boundaries of change and no-change classes. These example images are from the LEVIR building CD dataset~\cite{levid-cd}. \vspace{-5mm}}
    \label{fig:cluster_assump}
\end{figure}
\vspace{-3mm}
The cluster assumption is the key to successfully applying consistency regularization in semi-supervised learning. The \textit{cluster assumption} states that when low-density regions separate classes, we can efficiently utilize unlabeled data to refine the decision boundaries by consistency regularization. Following this notion, we can redefine the cluster assumption for the CD problem, where we can think of low-density regions as the regions with no-changes and high-density regions as change regions. Therefore, for CD, the cluster assumption holds if the boundary between low-density and high-density regions align with the boundary between change and no-change classes of the ground-truth change map. If the cluster assumption holds for CD, we can apply small random perturbations to each data point and make the predicted change map from the deep network to be consistent. As one can realize, we cannot guarantee such consistency when change and no-change class clusters overlap with each other, as applying a small perturbation could lead to altered class labels. Having formulated the cluster assumption for CD, let us now try to visualize it in both the input bi-temporal image domain and the hidden feature difference domain. To visualize the cluster assumption in the input bi-temporal image domain, we compute the average $L^2$ distance between a patch of size $15 \times 15$ with its immediately neighboring eight patches. In this case, we consider input bi-temporal image $I$ as the concatenation of pre-change $I_A$ and post-change image $I_B$ along their spectral dimension. In order to visualize the cluster assumption in the hidden feature difference domain, we compute the average square distance between the $512-$dimensional hidden feature difference vector and its eight immediately neighboring vectors. In Fig. \ref{fig:cluster_assump}, we visualize the average $L^2$ distance in the input bi-temporal image domain and hidden feature difference domain for two example images from the LEVIR dataset~\cite{levid-cd}. From fig. \ref{fig:cluster_assump} - (d) and (e), one can notice that the cluster assumption for CD does not hold in the input bi-temporal image domain since the low-density regions (in \textcolor{blue}{blue} color) do not align with the boundaries of the change-class. In contrast, the cluster assumption holds for the hidden feature difference domain where the change-class has a high average distance, corresponding to high-density regions.

Since, the cluster assumption for CD holds in hidden feature difference domain, we can now utilize consistency regulation to leverage the information from freely-available, unlabeled, bi-temporal remote-sensing images to enhance the performance of supervised CD.
\vspace{-4mm}
\subsection{Proposed semi-supervised CD approach}
\begin{figure}[!tb]
    \centering
    \label{fig:network}
    \includegraphics[trim=1.5cm 0cm 1.5cm 0cm, width=\linewidth]{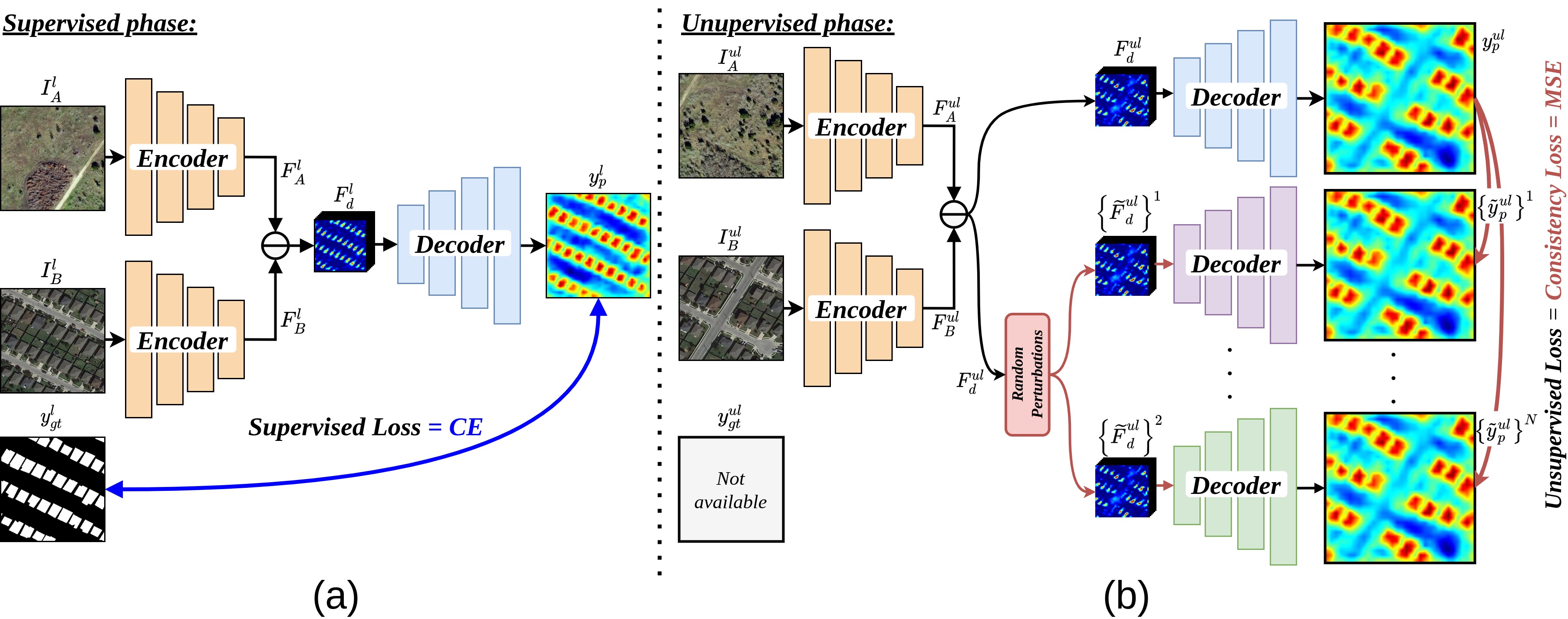}
    \caption{The proposed semi-supervised approach for CD. \textbf{(a)} Supervised phase. \textbf{(b)} Unsupervised phase. \vspace{-5mm}}
\end{figure}

\subsubsection{CD network architecture.} As shown in Fig. \ref{fig:network}, our CD network consists of three main components:
\begin{enumerate}
     \item \textit{An Encoder $f_{e}$} to extract hidden feature representations $F_A$ and $F_B$ of pre-change image $I_A$ and post-change image $I_B$,
     \item \textit{A feature difference module} to obtain the difference of hidden feature representations $F_d$ of pre-change and post-change images, and
     \item \textit{A decoder $f_d$} to predict the change map from hidden feature difference $F_d$.
 \end{enumerate}
\paragraph{The Encoder.} For the encoder, we use a  pre-trained ResNet50~\cite{he2016deep}. The output of the encoder is a 2048-dimensional feature cube with a spatial resolution of $H/4  \times W/4$, where $H$ and $W$ are the height and width of the input bi-temporal image $\{I_A, I_B\}$, respectively. In particular, we utilize the encoder in a Siamese network architecture~\cite{bromley1993signature}, where we share the weights of the encoder with pre-change $I_A$ and post-change $I_B$ image to obtain their hidden feature representations $F_A$ and $F_B$. We can mathematically formulate the above process as:
\setlength{\belowdisplayskip}{0pt} \setlength{\belowdisplayshortskip}{0pt}
\setlength{\abovedisplayskip}{0pt} \setlength{\abovedisplayshortskip}{0pt}
\begin{align}
    F_A &= f_e(I_A),\\
    F_B &= f_e(I_B).
\end{align}
\paragraph{The feature difference module.} Once we have the hidden feature representations $F_A$ and $F_B$ from the encoder for a given bi-temporal image $\{I_A, I_B\}$, we compute the absolute difference between $F_A$ and $F_B$ to obtain the difference in the hidden features, and subsequently process it through a Pyramid Pooling Module (PPM)~\cite{zhao2017pyramid} $f_{\text{PPM}}$ to effectively harvest changes in different scales. We denote the output from the above process as hidden feature difference $F_d$ of bi-temporal image $\{I_A, I_B\}$. We can mathematically express the process inside the feature difference module  as:
\begin{align}
    F_d &= f_{\text{PPM}} \left( \abs{F_A - F_B}_1 \right).
\end{align}
\paragraph{The decoder $f_d$.} The purpose of the decoder is to estimate the output change probability map $\hat{y}$ from the hidden feature difference $F_d$. We use series of sub-pixel convolutional upsampling modules \cite{shi2016real}, until we reach the spatial resolution of the input bi-temporal images $H \times W$. The process inside the decoder can mathematically express as:
\begin{equation}
    \hat{y} = f_d(F_d).
\end{equation}

We optimize the parameters of our CD model $f_{CD}(\cdot) = f_d(f_{\text{PPM}}(f_e(\cdot)))$ using the proposed semi-supervised technique which utilizes labeled as well as unlabeled bi-temporal images available for training. 

\subsubsection{Learning from labeled data (supervised phase).} In this phase, we optimize the parameters of our CD network $f_{CD}(\cdot)$ to predict only the \textit{changes of interest} implied by the labeled training data: $\mathcal{D}_{l} = \left\{ \{I_{A,i}^l, I_{B,i}^l\}, y_{i}^l \right\}_{i=1}^{N_l} $, where $\{I_{A,i}^l$, $I_{B,i}^l\}$ denotes the $i$-th bi-temporal image, $y_{i}^l$ is the corresponding ground-truth change mask, and $N_l$ is the size of the labeled dataset.  To this end, we utilize the Cross Entropy ($\texttt{CE}$) loss ~\cite{murphy2012machine} as the supervised loss $\mathcal{L}_{sup}$ which measures the performance of the predicted change probability map $\hat{y}_{i}^l = f_{CD}(I_{A,i}^l, I_{B,i}^l)$ compared to the actual change map $y^l_{i}$ as:
\begin{equation}
    \mathcal{L}_{sup} = \texttt{CE}(\hat{y}_{i}^l, y^l_{i}).
\end{equation}
The above process is graphically depicted in Fig. \ref{fig:cluster_assump} - (a).

\subsubsection{Learning from unlabeled data (unsupervised phase).} In this phase, we make use of unlabeled bi-temporal images $\mathcal{D}_{ul} = \left\{ I_{A,i}^{ul}, I_{B,i}^{ul} \right\}_{i=1}^{N_{ul}}$ in addition to the labeled data $\mathcal{D}_{l}$, where $\{ I_{A,i}^{ul}, I_{B,i}^{ul} \}$ is the $i$-th unlabeled bi-temporal image and $N_{ul}$ is the size of the unlabeled dataset which is generally assumed to be higher than the size of the labeled dataset (i.e., $N_{ul} >> N_l$). Now, our goal is to effectively utilize this readily-available unlabeled bi-temporal images to improve the performance of our change detection model $f_{CD}(\cdot)$ over the case where we have only labeled data. To archive this goal, we formulate an unsupervised loss $\mathcal{L}_{unsup}$ based on unlabeled data which provides an additional training signal to optimize the parameters of $f_{CD}(\cdot)$. Our unsupervised loss is based on the \textit{cluster assumption} in CD that we presented in Sec. \ref{sec:cluster_assumption}, where we enforce the predictions of our $f_{CD}(\cdot)$ to be consistent under small random perturbations applied on the deep feature difference map $F_d$ as shown in Fig. \ref{fig:cluster_assump}-(b).

Let's define  $\left\{ (\widetilde{F}_{d,i}^{ul})_{1}, \cdots, (\widetilde{F}_{d,i}^{ul})_{p}, \cdots, (\widetilde{F}_{d,i}^{ul})_{p=N_p} \right\}$ as the set of randomly perturbed versions of the hidden feature difference $F_{d,i}^{ul}$ of the $i$-th unlabeled bi-temporal image pair $\{I_{A,i}^{ul}, I_{B,i}^{ul}\}$. Next, we process $F_{d,i}^{ul}$ through our main decoder $f_d(\cdot)$ and obtain the predicted change probability map as:
\begin{equation}
    \hat{y}_i^{ul} = f_d\left(F_{d,i}^{ul}\right).
\end{equation}
Similarly, we process each perturbed version of the hidden feature difference map through a set of $N_p$ auxiliary decoders which are similar in design as $f_d(\cdot)$ and obtain their corresponding predictions $(\widetilde{y}_i^{ul})_p$ as:
\begin{equation}
    \left(\widetilde{y}_i^{ul}\right)_{p} = f_d^p\left( (\widetilde{F}_{d,i}^{ul})_{p} \right) \text{, where } p=1, \dots ,N_p.
\end{equation}
Next, we enforce $\left\{ \left(\widetilde{y}_i^{ul}\right)_{p} \right\}_{p=1}^{N_p}$ to be consistent with $\hat{y}_i^{ul}$ by defining the unsupervised loss $\mathcal{L}_{unsup}$  as follows:
\begin{equation}
    \mathcal{L}_{unsup} = \sum_{p=1}^{N_p} \mathbf{d} \left( \left(\widetilde{y}_i^{ul}\right)_{p}, \hat{y}_i^{ul} \right),
\end{equation}
where $\mathbf{d}(\cdot)$ is a distance metric that measures the dissimilarity between the predictions $\left(\widetilde{y}_i^{ul}\right)_{p}$ and $\hat{y}_i^{ul}$. For our experiments we use $\mathbf{d}(\cdot)$ as the mean squares error ($\texttt{MSE}$). 

\vspace{-5mm}
\subsubsection{Types of perturbations.} In our experiments, we use following perturbations:
\label{sec:types_of_pertb}
\begin{itemize}
    \item \textit{Random feature noise:} We generate a 3D noise tensor $\mathbf{N} \sim \mathcal{U}(-0.3, 0.3)$, then scale it according to the magnitude of values in $F_{d,i}^{ul}$, and add it to the hidden feature difference map to get the perturbed version: $\widetilde{F}_{d,i}^{ul} = F_{d,i}^{ul} + \mathbf{N} \odot F_{d,i}^{ul}$.
    \item \textit{Random feature drop:} We mask out 10\% to 40\% of the most needed regions in the feature difference map by first sampling a threshold $\gamma \sim \mathcal{U}(0.6, 0.9)$, then generating the mask $ \mathbf{M} =  \{ \overline{F_{d,i}^{ul}} <  \gamma \}_{\mathbf{1}}$ where $\overline{F_{d,i}^{ul}}$ is the normalized feature difference map $F_{d,i}^{ul}$ along the channel-dimension, and finally element-wise multiplying it with the obtained mask to get the perturbed feature difference map $\widetilde{F}_{d,i}^{ul} = \mathbf{M} \odot \widetilde{F}_{d,i}^{ul}$.
    \item \textit{Guided Feature Cutout:} To reduce the CD network's dependency on a specific part of the feature difference map, we randomly zero-out a random-crop from the feature difference map based on the predicted change map $\hat{y}^{ul}_i$ to get perturbed feature difference map $\widetilde{F}_{d,i}^{ul}$.
    \item \textit{Content and object masking:} Based on the assumption that the CD network's output should be invariant to the change or no-change classes, we create two perturb versions of the hidden feature difference map by masking out change region in the hidden feature difference map by change-mask $\mathbf{M}_c$ or no-change region in the hidden feature difference map by a nochange-mask $\mathbf{M}_{nc}=1-\mathbf{M}_c$. 
    \item{\textit{Feature-VAT}~\cite{miyato2018virtual}:} To make our feature difference map isotropically smooth around each data vector, we apply adversarial perturbation to it in the direction where it will alter most as: $\widetilde{F}_{d,i}^{ul} = p_{adv} + F_{d,i}^{ul}$. 
\end{itemize}
\vspace{-5mm}
\subsubsection{Final loss function.} We define our final loss function $\mathcal{L}$ as follows,
\begin{equation}
    \mathcal{L} = \mathcal{L}_{sup} + \lambda (t) \mathcal{L}_{unsup},
\end{equation}
where $\lambda(t)$ is a regularization constant which is a function of the training iteration number $t$. In particular, we vary $\lambda(t)$ from $0$ to $1$ like the Gaussian function for the first $T$ number of iterations and keep it constant at $1$ for the remaining iterations $t>T$.
\vspace{-5mm}
\section{Experimental setup}
\subsubsection{Datasets.} 
\vspace{-4mm}
We use two publicly available, widely-used CD datasets for our experiments, namely LEVIR~\cite{levid-cd} and WHU~\cite{whu-cd}. The LEVIR and WHU are building CD datasets. Following previous works on supervised CD~\cite{bit,changeformer}, we create non-overlapping patches of size $256 \times 256$ for the training, for both the datasets while utilizing their default \texttt{train}, \texttt{val}, and \texttt{test} sets. In the semi-supervised setting, we randomly select the required number of labeled training data for each case from the original \texttt{train}-set and consider the rest of the images in the \texttt{train}-set as unlabeled data by discarding the corresponding change labels. We optimize CD models on \texttt{val}-set and report the results on the \texttt{test}-set.
\vspace{-5mm}
\subsubsection{Performance metrics.} Following~\cite{bit,changeformer}, we use Intersection over Union $(IoU)$ as the primary performance metric to quantify the CD model's performance. Since a better CD model should predict both change and no-change regions correctly, we report $IoU$ metric w.r.t. change-class $(IoU^{c})$ along with the overall pixel accuracy (\texttt{OA}) which measures the proportion of correctly predicted pixels.
\vspace{-5mm}
\subsubsection{Implementation details.} We implemented our model in \texttt{PyTorch} and trained using an \texttt{NVIDIA Quadro RTX 8000} GPU. During training, we applied data augmentation through random flip, random re-scale ($0.8-1.2$), random crop, Gaussian blur, and random color jittering. The learning rate is initially set equal to $0.01$. We trained the model for 80 epochs, and  used a batch size of $8$ for both labeled and unlabeled datasets. The code and pre-trained models will be publicly made available after the review process. 

Please refer supplementary material for more details on the experimental setup.

\vspace{-4mm}
\section{Results and discussion}
\vspace{-3mm}
In order to demonstrate the superiority of our semi-supervised CD method, we perform both quantitative and qualitative comparisons with SOTA methods. As we discussed in Sec. ~\ref{sec:sota_semi}, SemiCDNet~\cite{SemiCDNet} is the current and the only SOTA method available for semi-supervised CD. However, to make the analysis more comprehensive, we also compare our proposed method with SOTA semi-supervised segmentation methods, namely AdvNet~\cite{advnet} and s4GAN~\cite{s4GAN} by properly re-implementing them for the CD task.

\begin{table}[tb]
	\centering
	\tiny
	\caption{The average quantitative metrics of different CD methods on LEVIR$\rightarrow$LEVIR and WHU$\rightarrow$WHU with the percentage of labeled data.}
	\begin{tabular}{p{18mm}p{5mm}p{5mm}cp{5mm}p{5mm}cp{5mm}p{5mm}cp{5mm}p{5mm}c|p{5mm}p{5mm}cp{5mm}p{5mm}cp{5mm}p{5mm}cp{5mm}p{5mm}} %
		\toprule
		 & \multicolumn{11}{c}{LEVIR(\% labeled)$\rightarrow$LEVIR} && \multicolumn{11}{c}{WHU(\% labeled)$\rightarrow$WHU} \\
		\cmidrule{2-12} \cmidrule{14-24} 
		\multirow{2}{*}{\parbox[c]{.2\linewidth}{Method}} & \multicolumn{2}{c}{5\%} & & \multicolumn{2}{c}{10\%} & & \multicolumn{2}{c}{20\%} & & \multicolumn{2}{c}{40\%} && \multicolumn{2}{c}{5\%} && \multicolumn{2}{c}{10\%} && \multicolumn{2}{c}{20\%} && \multicolumn{2}{c}{40\%}\\ 
		\cmidrule{2-3} \cmidrule{5-6} \cmidrule{8-9} \cmidrule{11-12} \cmidrule{14-15} \cmidrule{17-18} \cmidrule{20-21} \cmidrule{23-24}
		
		& {$IoU^c$} & {OA} && {$IoU^c$} & {OA} & & {$IoU^c$} & {OA} &&{$IoU^c$} & {OA} && {$IoU^c$} & {OA} && {$IoU^c$} & {OA} && {$IoU^c$} & {OA} && {$IoU^c$} & {OA}\\
		\midrule
		Sup. only   &   61.0 & 97.60 && 
		                66.8 & 98.13 && 
		                72.3 & 98.44 && 
		                74.9 & 98.60 && 
		                50.0 & 97.48 && 
		                55.7 & 97.53 && 
		                65.4 & 98.20 && 
		                76.1 & 98.94 \\ 
		AdvNet\cite{advnet}& 66.1 & 98.08 && 
		                72.3 & 98.45 && 
		                74.6 & 98.58 && 
		                75.0 & 98.60 && 
		                55.1 & 97.90 &&  
		                61.6 & 98.11 && 
		                73.8 & 98.80 && 
		                76.6 & 98.94\\ 
		s4GAN\cite{s4GAN}& 64.0 & 97.89 && 
		                67.0 & 98.11 && 
		                73.4 & 98.51 &&
		                75.4 & 98.62 &&
		                18.3 & 96.69 && 
		                62.6 & 98.15 && 
		                70.8 & 98.60 &&
		                76.4 & 98.96\\
		SemiCDNet\cite{SemiCDNet} & 67.6 & 98.17 && 
		                71.5 & 98.42 && 
		                74.3 & 98.58 && 
		                75.5 & 98.63 && 
		                51.7 & 97.71 && 
		                62.0 & 98.16 && 
		                66.7 & 98.28 && 
		                75.9 & 98.93\\ 
		Ours        &   \textcolor{red}{72.5} & \textcolor{red}{98.47} && 
		                \textcolor{red}{75.5} & \textcolor{red}{98.63} &&  
		                \textcolor{red}{76.2} & \textcolor{red}{98.68} && 
		                \textcolor{red}{77.2} & \textcolor{red}{98.72} && 
		                \textcolor{red}{65.8} & \textcolor{red}{98.37} && 
		                \textcolor{red}{68.1} & \textcolor{red}{98.47} && 
		                \textcolor{red}{74.8} & \textcolor{red}{98.84} && 
		                \textcolor{red}{77.2} & \textcolor{red}{98.96}\\ 
		\hline
		Oracle & \multicolumn{11}{c}{$ IoU^c$=\textcolor{red}{\bf 77.9} and OA=\textcolor{red}{\bf 98.77}} && \multicolumn{11}{c}{$IoU^c$=\textcolor{red}{\bf 85.5} and OA=\textcolor{red}{\bf 99.38}}\\
		\bottomrule
	\end{tabular}
	\normalsize
	\label{tab:same_dataset}
\end{table}

\vspace{-4mm}
\subsection{Experiments within the same dataset}
\label{sec:same_dataset}
In this experiment, we compare the CD performance of our method with SOTA methods where training and testing were performed on the same dataset (denoted by LEVIR$\rightarrow$LEVIR and WHU$\rightarrow$WHU\footnote[1]{Specifically, $X$$\rightarrow$$Y$ denotes training a model on the  $\texttt{train}$-set of dataset $X$ and testing it on the $\texttt{test}$-set of dataset $Y$.}). Table \ref{tab:same_dataset} summarizes the $IoU^c$ and $OA$ for LEVIR$\rightarrow$LEVIR and WHU$\rightarrow$WHU for different \% of labeled data. The key takeaways from Tab. \ref{tab:same_dataset} can be summarized as follows: \textbf{(1)} Increasing the \% of labeled data in training results in increased CD performance for all methods, showing that ``one can achieve better results when there is access to more labeled data.'' \textbf{(2)} However, for the same amount of labeled data, all semi-supervised CD methods (i.e., AdvNet~\cite{advnet}, s4GAN~\cite{s4GAN}, and SemiCDNet~\cite{SemiCDNet}) give better results than supervised CD, empirically validating that ``CD can benefit from the freely available, unlabeled, remote-sensing data''. \textbf{(3)} Notably, our semi-supervised CD method outperforms the SOTA SemiCDNet method~\cite{SemiCDNet} by a significant margin, empirically demonstrating that our method can leverage information from unlabeled data more efficiently than its counterparts. More importantly, when the amount of labeled data is very less (5\% and 10\% of cases), our method significantly outperforms the SOTA method by at least a 5\% increase in the IoU metric.

\begin{table}[tb]
	\centering
	\tiny
	\caption{The average quantitative metrics of different CD methods for LEVIR $\rightarrow$ WHU and WHU $\rightarrow$ LEVIR with the percentage of labeled data.}
	\begin{tabular}{p{18mm}p{5mm}p{5mm}cp{5mm}p{5mm}cp{5mm}p{5mm}cp{5mm}p{5mm}c|p{5mm}p{5mm}cp{5mm}p{5mm}cp{5mm}p{5mm}cp{5mm}p{5mm}} %
		\toprule
		 & \multicolumn{11}{c}{LEVIR (\% labeled) $\rightarrow$ WHU} && \multicolumn{11}{c}{WHU (\% labeled) $\rightarrow$ LEVIR} \\
		\cmidrule{2-12} \cmidrule{14-24} 
		\multirow{2}{*}{\parbox[c]{.2\linewidth}{Method}} & \multicolumn{2}{c}{5\%} & & \multicolumn{2}{c}{10\%} & & \multicolumn{2}{c}{20\%} & & \multicolumn{2}{c}{40\%} && \multicolumn{2}{c}{5\%} && \multicolumn{2}{c}{10\%} && \multicolumn{2}{c}{20\%} && \multicolumn{2}{c}{40\%}\\ 
		\cmidrule{2-3} \cmidrule{5-6} \cmidrule{8-9} \cmidrule{11-12} \cmidrule{14-15} \cmidrule{17-18} \cmidrule{20-21} \cmidrule{23-24}
		
		& {$IoU^c$} & {OA} && {$IoU^c$} & {OA} & & {$IoU^c$} & {OA} &&{$IoU^c$} & {OA} && {$IoU^c$} & {OA} && {$IoU^c$} & {OA} && {$IoU^c$} & {OA} && {$IoU^c$} & {OA}\\
		\midrule
		Sup. only   &   24.5 & 95.90 && 
		                32.2 & 96.70 && 
		                38.7 & 96.59 && 
		                47.2 & 97.12 && 
		                3.7 & 94.68 && 
		                10.9 & 93.60 && 
		                9.2 & 92.81 && 
		                6.6 & 93.95\\ 
		AdvNet\cite{advnet}&27.2 & 96.37 && 
		                33.2 & 96.56 && 
		                32.8 & 96.97 && 
		                43.1 & 97.33 && 
		                4.3 & 94.44 && 
		                11.0 & 93.56 && 
		                11.6 & 92.57 && 
		                7.3 & 93.85\\ 
		s4GAN\cite{s4GAN}&  31.2 & 96.33 && 
		                31.7 & 95.65 && 
		                41.9 & 96.95 &&
		                47.9 & 97.63 &&
		                1.6 & 94.94 && 
		                7.8 & 94.30 && 
		                8.7 & 93.22 &&
		                2.2 & 94.29\\
		SemiCDNet\cite{SemiCDNet} & 28.4 & 96.41 && 
		                32.5 & 96.93 && 
		                39.5 & 97.20 && 
		                48.3 & 97.53 && 
		                3.4 & 94.68 && 
		                8.9 & 94.03 && 
		                9.7 & 92.16 && 
		                5.3 & 94.60\\ 
		Ours        &   \textcolor{red}{35.4} & \textcolor{red}{96.40} && 
		                \textcolor{red}{39.3} & \textcolor{red}{96.76} && 
		                \textcolor{red}{42.1} & \textcolor{red}{96.99} && 
		                \textcolor{red}{51.5} & \textcolor{red}{97.72} && 
		                \textcolor{red}{7.8} & \textcolor{red}{94.72} && 
		                \textcolor{red}{11.7} & \textcolor{red}{94.48} && 
		                \textcolor{red}{10.6} & \textcolor{red}{93.82} && 
		                \textcolor{red}{12.3} & \textcolor{red}{94.92}\\ 
		\bottomrule
	\end{tabular}
	\normalsize
	\label{tab:cross_dataset}
\end{table}
\begin{figure}[!htb]
    \centering
    \includegraphics[width=\linewidth]{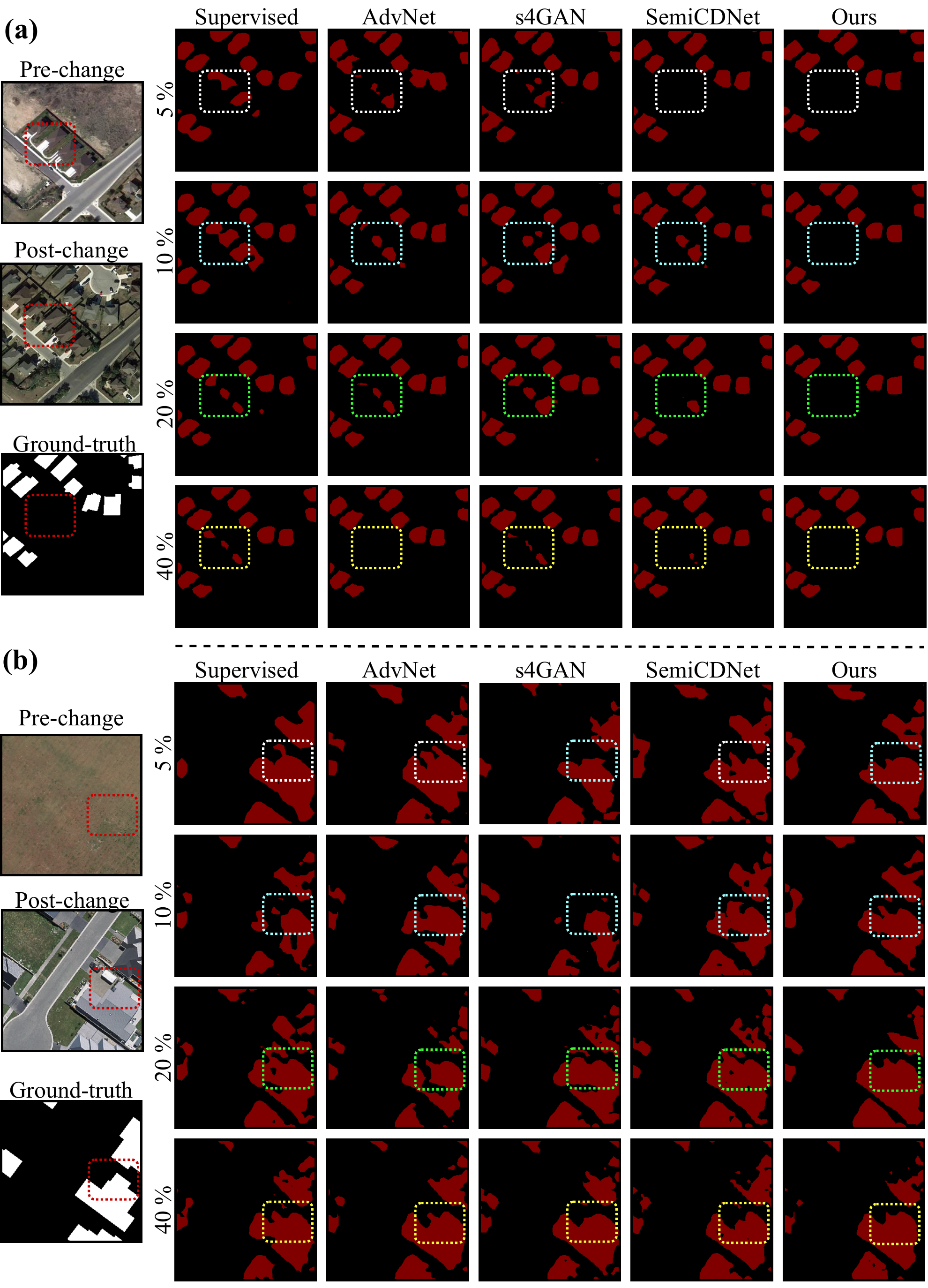}
    \caption{Qualitative results of different CD methods with \% of labeled data for \textbf{(a)} testing on the same dataset: LEVIR $\rightarrow$ LEVIR and \textbf{(b)} testing on a different dataset: LEVIR $\rightarrow$ WHU. See supplementary material for additional qualitative results.}
    \label{fig:qualitative_res}
\end{figure}

\vspace{-4mm}
\subsection{Model generalizability/transferability}
\label{sec:cross_dataset}
In this experiment, our goal is to test how well our model generalizes to other datasets compared to the SOTA methods. To analyze this, we take the models trained in the previous Section \ref{sec:same_dataset} and test them on another dataset for the same task. In particular, models trained on LEVIR are tested on WHU and vice versa. Note that both LEVIR and WHU are building CD datasets; therefore, if the model is well generalized, then it should give better CD results not only for the one we trained but also on the other datasets. Table \ref{tab:cross_dataset} summarizes the results from LEVIR $\rightarrow$ WHU and WHU $\rightarrow$ LEVIR with different \% of labeled data used in training. The key takeaways from Tab. \ref{tab:cross_dataset} can be summarized as follows: \textbf{(1) } Semi-supervised CD models produce better CD results than the supervised model, confirming the generally accepted belief that ``supervised methods are generally less generalized.'' \textbf{(2)} In particular, the proposed method archives outstanding CD results in both datasets, confirming that our method is much more generalized than the SOTA methods. Especially in the case where the labeled data is less  (i.e., 5\% and 10\%), the improvement of the proposed method over the SOTA is more than 22\%, which demonstrates a strong capability of generalization. This could be because the proposed approach is trained to produce consistent results even under small perturbations that appear on hidden feature difference maps - making it more robust and generalized to slight variations appearing on the hidden feature differences when testing on another dataset. \textbf{(3)} The reason why SOTA methods are less generalized compared to our method can be explained as follows. As we explained in Section \ref{sec:sota_semi}, the SOTA method uses adversarial loss to exploit information from unlabeled data where it assumes that the change maps for both labeled and unlabeled data follow the same distribution.  This forces the CD network to predict change maps similar to the ones seen during training, leading to poor generalization when the target data follows a different distribution. In contrast, we do not assume such an assumption and utilize the consistency training to leverage the information from unlabeled data, making it a more generalized CD method.

\begin{table}[tb]
	\centering
	\scriptsize
	\caption{The average quantitative metrics of different CD methods for \{LEVIR(sup.\%), WHU (unsup.)\} $\rightarrow$ LEVIR with the percentage of labeled data.\vspace{-2mm}}
	\begin{tabular}{lllcllcllcllcll} %
		\toprule
		 & \multicolumn{14}{c}{\{LEVIR(sup.\%),WHU(unsup.)\}$\rightarrow$LEVIR} \\
		\cmidrule{2-15}
		\multirow{2}{*}{\parbox[c]{.2\linewidth}{Method}} & \multicolumn{2}{c}{5\%} & & \multicolumn{2}{c}{10\%} & & \multicolumn{2}{c}{20\%} & & \multicolumn{2}{c}{40\%} & & \multicolumn{2}{c}{100\%}\\ 
		\cmidrule{2-3} \cmidrule{5-6} \cmidrule{8-9} \cmidrule{11-12} \cmidrule{14-15}
		
		& {$IoU^c$} & {OA} && {$IoU^c$} & {OA} & & {$IoU^c$} & {OA} &&{$IoU^c$} & {OA} &&{$IoU^c$} & {OA}\\
		\midrule
		Sup. only   &   61.0 & 97.60 && 
		                66.8 & 98.13 && 
		                72.3 & 98.44 && 
		                74.9 & 98.60 && 
		                77.9 & 98.77\\ 
		AdvNet\cite{advnet}&65.2 & 98.10 && 
		                71.6 & 98.42 && 
		                75.0 & 98.63 && 
		                76.5 & 98.71 && 
		                77.4 & 98.78\\ 
		s4GAN\cite{s4GAN}&66.7 & 98.12 && 
		                68.4 & 98.26 && 
		                73.4 & 98.54 && 
		                75.3 & 98.65 && 
		                75.8 & 98.63\\ 
		SemiCDNet\cite{SemiCDNet} & 67.6 & 98.19 && 
		                71.9 & 98.46 && 
		                74.9 & 98.62 && 
		                76.4 & 98.70 && 
		                77.9 & 98.80\\ 
		Ours        &   \textcolor{red}{71.4} & \textcolor{red}{98.40} && 
		                \textcolor{red}{74.6} & \textcolor{red}{98.58} && 
		                \textcolor{red}{76.4} & \textcolor{red}{98.69} && 
		                \textcolor{red}{77.9} & \textcolor{red}{98.77} &&
		                \textcolor{red}{78.6} & \textcolor{red}{98.82}\\ 
		\bottomrule
	\end{tabular}
	
	\normalsize
	\label{tab:diferent_unl}
\end{table}
\begin{figure}[tb]
    \centering
    \includegraphics[width=\linewidth]{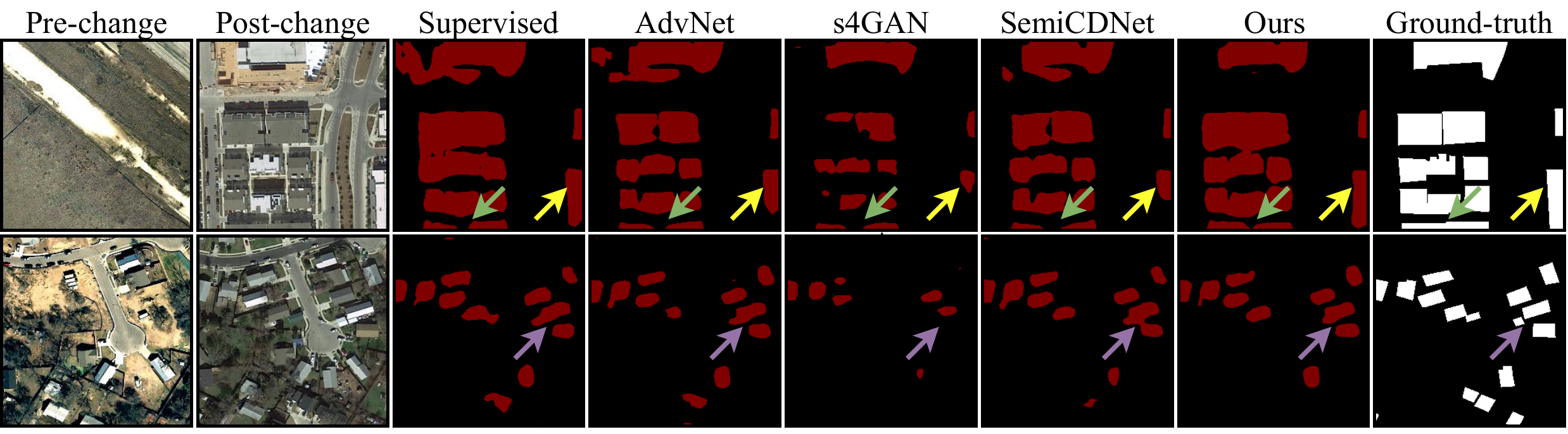}
    \caption{The qualitative results for \{LEVIR(100\%), WHU (unsup.)\} $\rightarrow$LEVIR.\vspace{-5mm}}
    \label{fig:qual_cross_data}
\end{figure}
\vspace{-3mm}
\subsection{Cross-dataset experiments}
\vspace{-3mm}
``Is it possible to improve CD results on one dataset by incorporating unlabeled remote sensing data from a different dataset?". To answer this question, in this experiment, we incorporate two datasets into the semi-supervised training process in which LEVIR is utilized as the labeled dataset and WHU is considered as the unlabeled dataset. We denote this scenario as: \{LEVIR(sup.\%), WHU(unsup.)\} $\rightarrow$ LEVIR. The quantitative results of this experiments are summarized in Tab. \ref{tab:diferent_unl}. From Tab. \ref{tab:diferent_unl}, it is clear that by incorporating unlabeled data even from another dataset can generally improve the CD results. More importantly, we can see that the proposed method outperforms the SOTA method SemiCDNet~\cite{SemiCDNet} by a significant margin in all cases, further demonstrating the higher capability of our method in leveraging the information from unlabeled data. Furthermore, we also present a qualitative comparison of our method with the SOTA methods in Fig. \ref{fig:qual_cross_data} for the scenario: \{LEVIR(100\%), WHU(unsup.) \}$\rightarrow$LEVIR. As highlighted in the figure by different colors of arrow heads, our method is able to predict more difficult changes accurately than all the methods under consideration.    

\section{Ablation study}
\begin{figure}[tb]
\begin{floatrow}
\capbtabbox{
  \scriptsize
  \begin{tabular}{c|l|l|l}
            \toprule
            Case & Configuration &  $IoU^c$ & $OA$\\
            \hline
            I   & Baseline (i.e., Sup.)& 66.8 & 98.13 \\
            \hline
            II  & Sup.+FN               & 74.7 & 98.55\\
            III & Sup.+FN+FD               & 74.8 & 98.60\\
            IV  & Sup.+FN+FD+GFC           & 74.9 & 98.60\\
            V   & Sup.+FN+FD+GFC+C\&O      & 75.1 & 98.61\\
            \multirow{ 2}{*}{VI}  & Sup.+FN+FD+GFC+C\&O & \multirow{ 2}{*}{\textcolor{red}{75.5}} & \multirow{ 2}{*}{\textcolor{red}{98.63}}\\
            &\hspace{15pt}+F-VAT (\textbf{final model}) &\\
            \bottomrule
        \end{tabular}
}{%
  \caption{How different types of perturbations contributes to proposed semi-supervised CD.\vspace{-4mm}}%
  \label{tab:abl_pertub}
}
\ffigbox{%
  \includegraphics[width=\linewidth]{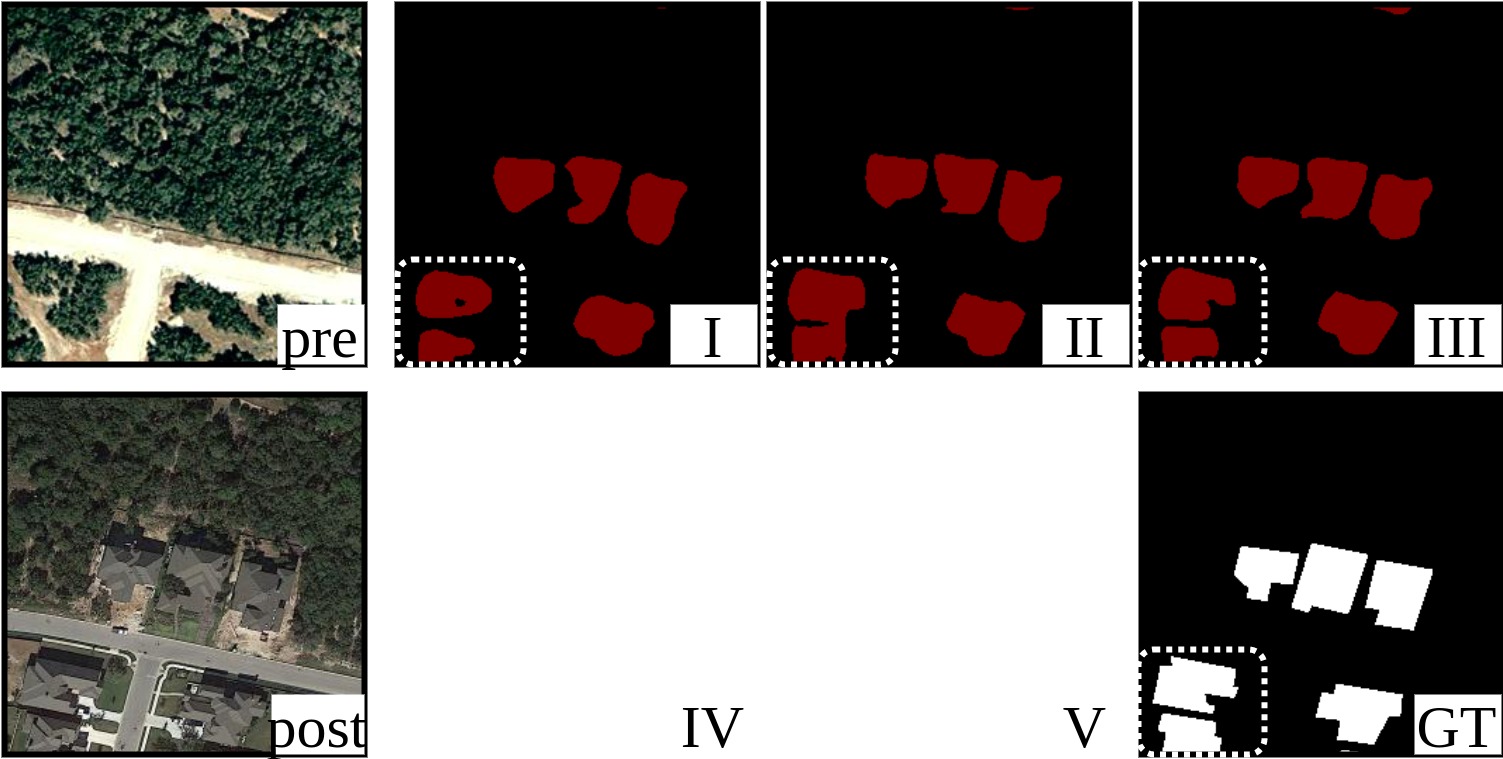}
}{%
  \caption{The qualitative improvements of predicted change maps with different types of perturbations.\vspace{-4mm}}%
  \label{fig:abl_pertub}
}
\end{floatrow}
\end{figure}

\vspace{-1mm}
In this ablation study, we demonstrate how different perturbations that we discussed in Sec. \ref{sec:types_of_pertb} contribute to the semi-supervised learning process. To demonstrate this, we conduct an experiment on LEVIR with 10\% labeled samples (i.e., LEVIR (10\%) $\rightarrow$ LEVIR). Starting from no-perturbations (i.e., supervised case) we progressively incorporate random Feature Noise (FN), random Feature Drop (FD), Guided Feature Cutout (GFC), Context \& Object masking (C\&T), and Feature-VAT (F-VAT) into the semi-supervised learning process, and the corresponding quantitative and qualitative results are summarized in Tab. \ref{tab:abl_pertub} and Fig. \ref{fig:abl_pertub}, respectively. As we can see both qualitatively and quantitatively, adding different types of random perturbations into the hidden feature difference map and then making the resulting change prediction map from the deep-network to be consistent results in significant improvement in the CD, while utilizing the same amount of annotated training images. 

\section{Conclusion}
\vspace{-1mm}
In this paper, we proposed a novel semi-supervised CD method that can leverage information from unlabeled RS data to improve the performance of CD by enforcing the predicted change probability map from the CD network to be consistent under different random perturbations applied on the hidden feature difference map. The proposed method can almost reach the supervised results with limited access to the annotated samples while outperforming the existing SOTA semi-supervised CD method (i.e., SemiCDNet) by a significant margin. Furthermore, experiments show that the proposed method has better generalizability and transferability compared to the SOTA semi-supervised CD methods. Since the proposed method does not assume any assumptions about the distribution for unlabeled change mask during the unsupervised training unlike GAN-based methods, it can also produce relatively better CD results when it is testing on another dataset that haven't been seen during training or when combining unlabeled data from completely different dataset to the semi-supervised learning process.

\clearpage
\bibliographystyle{splncs04}
\bibliography{egbib}
\end{document}


\pagestyle{headings}
\mainmatter

\title{Supplementary Materials for Revisiting Consistency Regularization for Semi-supervised Change Detection in Remote Sensing Images}

\titlerunning{Abbreviated paper title}
%
\author{Wele Gedara Chaminda Bandara \and Vishal M. Patel}
\authorrunning{W.G.C. Bandara et al.}
%
\institute{Johns Hopkins University, USA\\
\email{\{wbandar1, vpatel36\}@jhu.edu}}

\maketitle

\section{``\textit{Cluster assumption}'' in CD}
\begin{figure}[tbh]
    \centering
    \includegraphics[width=0.7\linewidth]{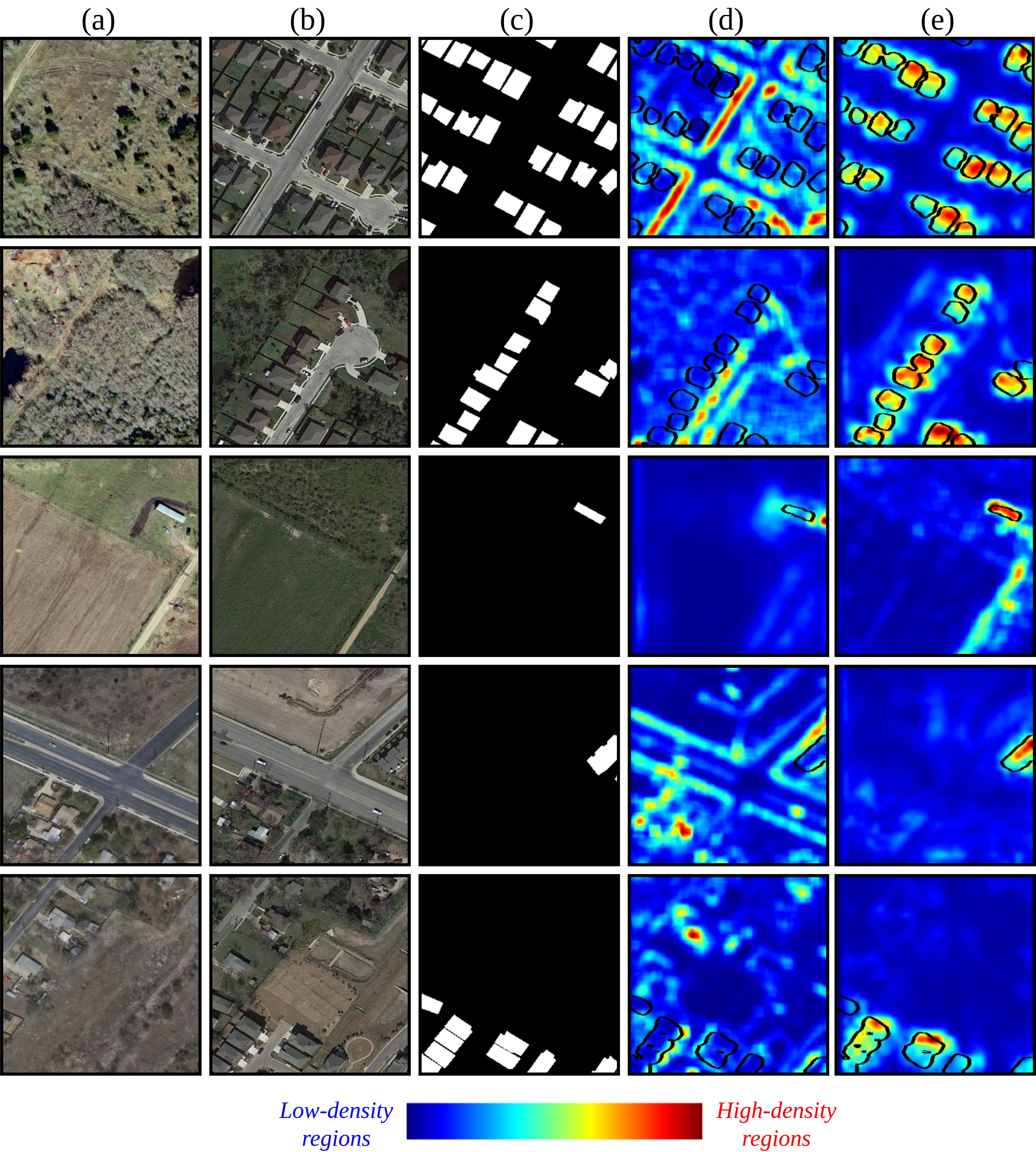}
    \caption{Additional examples to demonstrate the \textit{cluster assumption} in CD. \textbf{(a)} Pre-change image. \textbf{(b)} Post-change image. \textbf{(c)} Ground-truth change mask. \textbf{(d)} Cluster assumption in bi-temporal image domain. \textbf{(e)} Cluster assumption in hidden feature difference domain.}
    \label{fig_sup:cluster}
\end{figure}
Figure \ref{fig_sup:cluster} shows additional examples to demonstrate the cluster assumption in CD. From these images, we can also see that the cluster assumption does not hold in the bi-temporal image domain but rather holds in the deep feature difference domain. Therefore, we can leverage the information from freely-available, unlabeled, multi-temporal remote sensing images through consistency regularization by making the network's predicted change probability map consistent even under the small random perturbations applied on the hidden feature difference map. 

\clearpage
\section{Additional qualitative results for LEVIR$\rightarrow$LEVIR}

In this section, we provide additional qualitative results on LEVIR-CD dataset.  

\subsection{Additional qualitative examples for LEVIR(5\%)$\rightarrow$LEVIR}
\begin{figure}[!h]
    \centering
    \includegraphics[width=\linewidth]{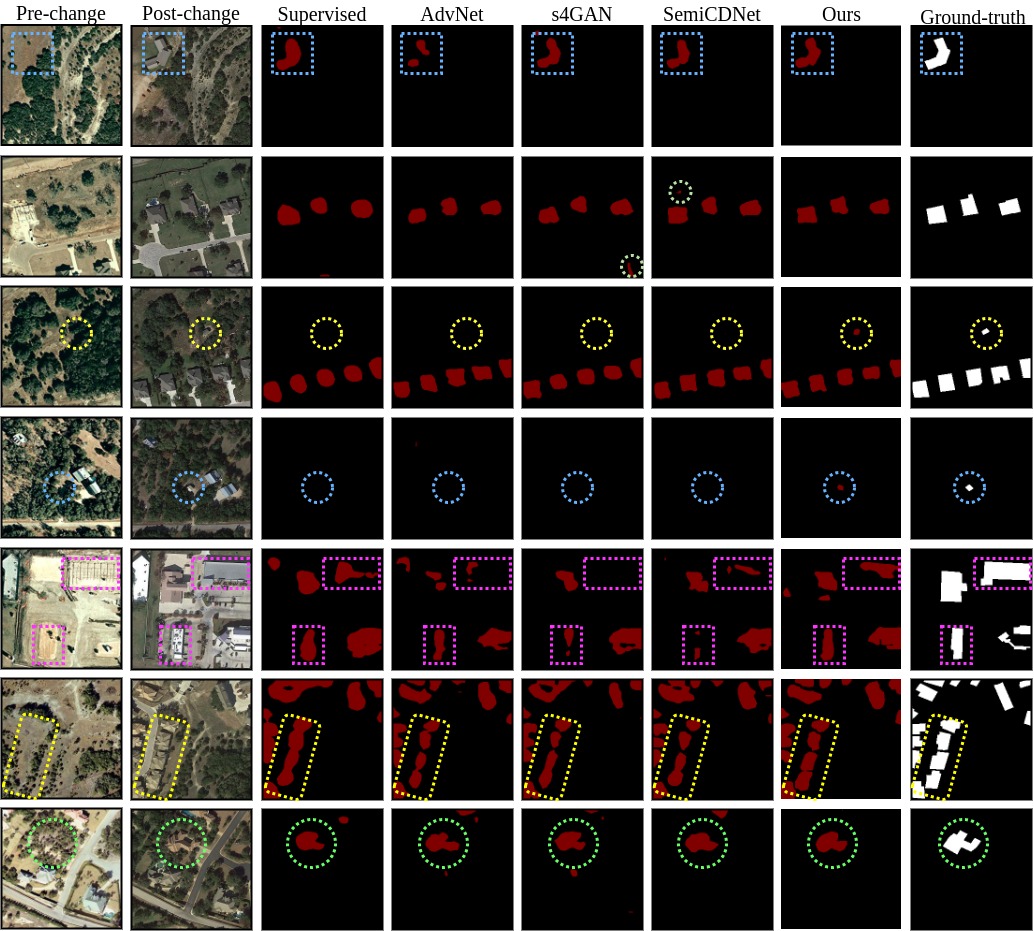}
    \caption{Additional qualitative results for LEVIR(5\%)$\rightarrow$LEVIR.}
    \label{fig:levir5-levir}
\end{figure}

Figure \ref{fig:levir5-levir} shows additional qualitative examples for LEVIR(5\%)$\rightarrow$LEVIR. From these examples also we can see that proposed semi-supervised approach generates much better change maps than the SOTA methods, further demonstrating its effectiveness of leveraging information from freely-available, unlabeled, remote-sensing data to improve the CD performance.

\clearpage
\subsection{Additional qualitative examples for LEVIR(10\%)$\rightarrow$LEVIR}
\begin{figure}[!h]
    \centering
    \includegraphics[width=\linewidth]{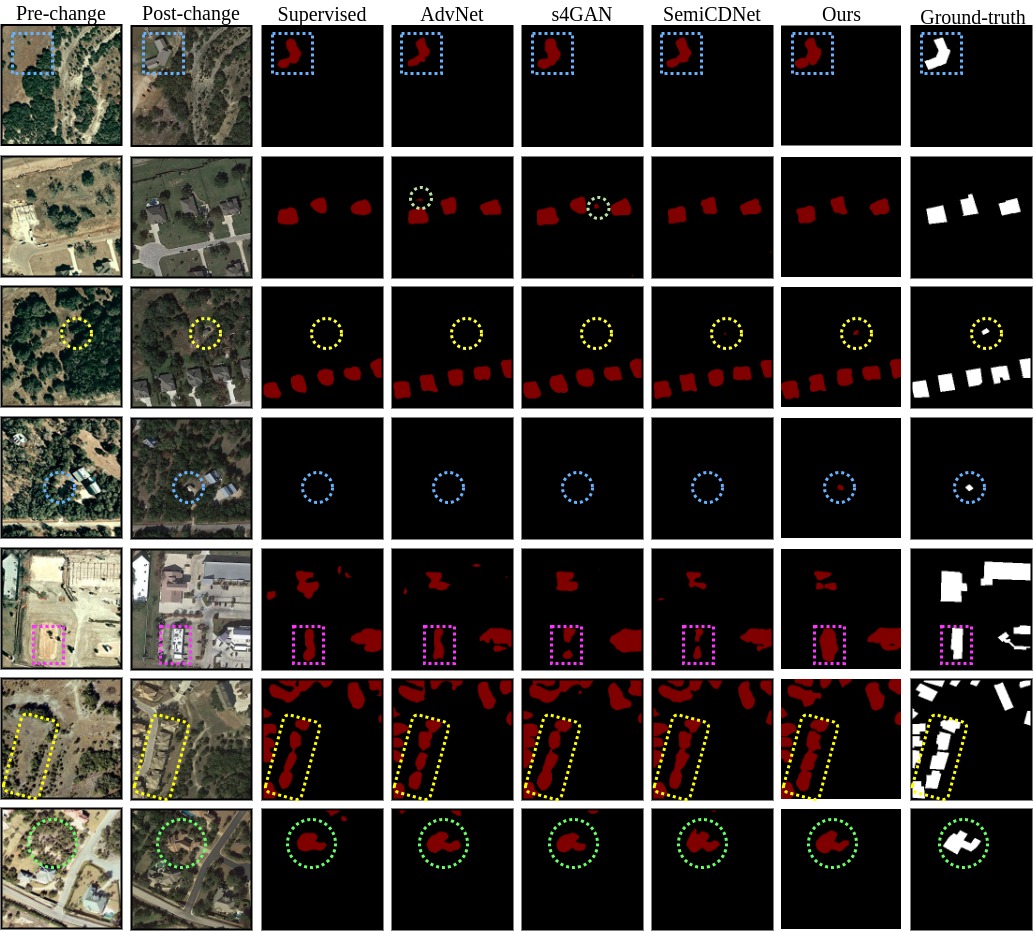}
    \caption{Additional qualitative results for LEVIR(10\%)$\rightarrow$LEVIR.}
    \label{fig:levir10-levir}
\end{figure}
Figure \ref{fig:levir10-levir} shows additional qualitative examples for LEVIR(10\%)$\rightarrow$LEVIR. Similar to Fig. \ref{fig:levir5-levir}, from these examples as well, we can see that the proposed semi-supervised approach generates much better change maps than SOTA methods, further demonstrating its effectiveness in leveraging freely-available, unlabeled remote-sensing images to improve CD performance.

\clearpage
\subsection{Additional qualitative examples for LEVIR(20\%)$\rightarrow$LEVIR}
\begin{figure}[!h]
    \centering
    \includegraphics[width=\linewidth]{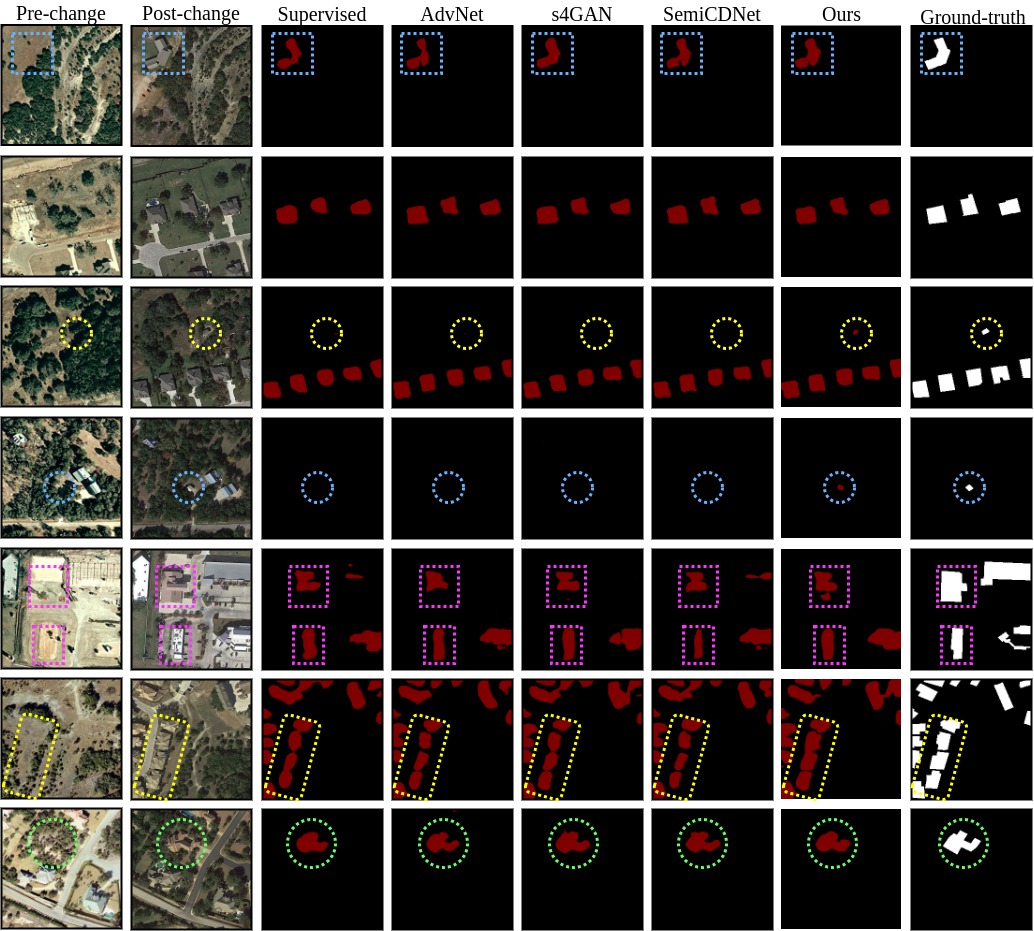}
    \caption{Additional qualitative results for LEVIR(20\%)$\rightarrow$LEVIR.}
    \label{fig:levir20-levir}
\end{figure}
Figure \ref{fig:levir20-levir} shows additional qualitative examples for LEVIR(20\%)$\rightarrow$LEVIR. Similar to Fig. \ref{fig:levir5-levir} and \ref{fig:levir10-levir},  from these examples as well, we can see that the proposed semi-supervised approach generates much better change maps than SOTA methods, further demonstrating its effectiveness in leveraging freely-available, unlabeled remote-sensing images to improve CD performance.

\clearpage
\subsection{Additional qualitative examples for LEVIR(40\%)$\rightarrow$LEVIR}
\begin{figure}[!h]
    \centering
    \includegraphics[width=\linewidth]{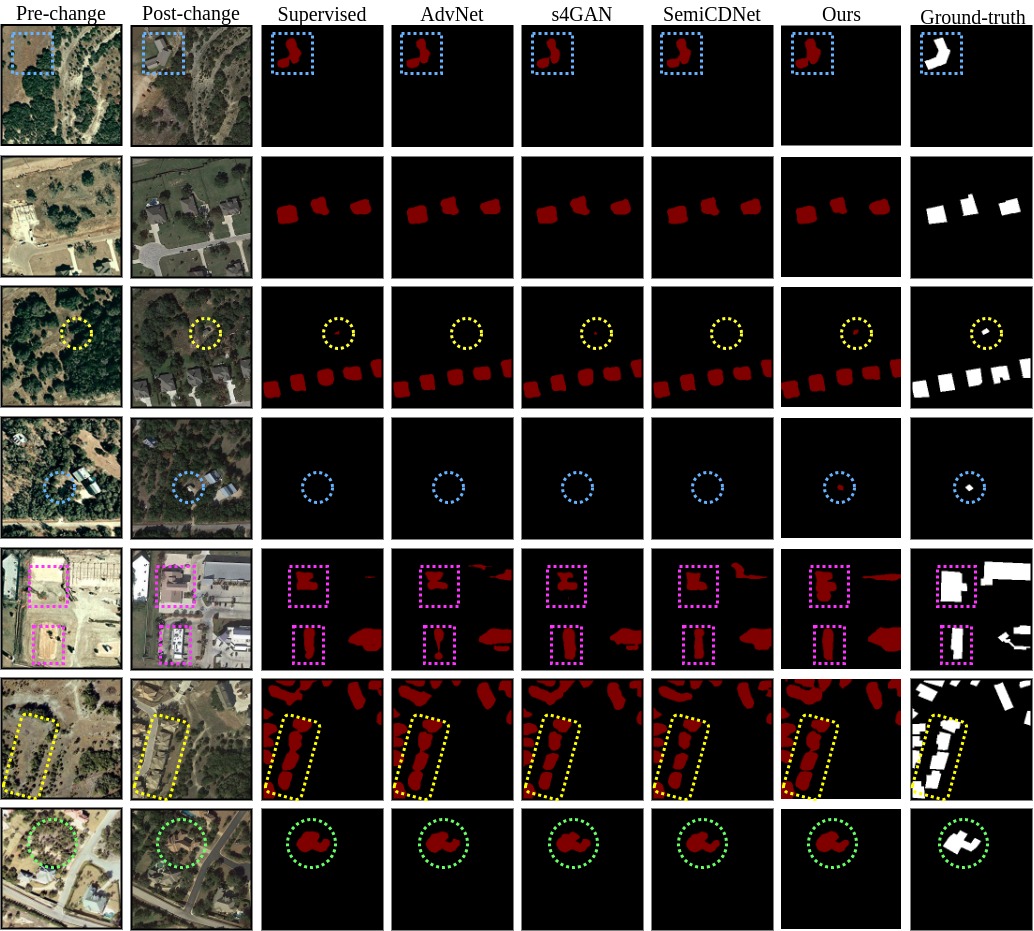}
    \caption{Additional qualitative results for LEVIR(40\%)$\rightarrow$LEVIR.}
    \label{fig:levir40-levir}
\end{figure}
Figure \ref{fig:levir40-levir} shows additional qualitative examples for LEVIR(40\%)$\rightarrow$LEVIR. Similar to Fig. \ref{fig:levir5-levir}, \ref{fig:levir10-levir}, and \ref{fig:levir20-levir},  from these examples as well, we can see that the proposed semi-supervised approach generates much better change maps than SOTA methods, further demonstrating its effectiveness in leveraging freely-available, unlabeled remote-sensing images to improve CD performance.

\clearpage
\section{Additional qualitative results for WHU$\rightarrow$WHU}
In this section, we provide additional qualitative results on WHU-CD dataset.

\subsection{Additional qualitative examples for WHU(5\%)$\rightarrow$WHU}
\begin{figure}[!h]
    \centering
    \includegraphics[width=\linewidth]{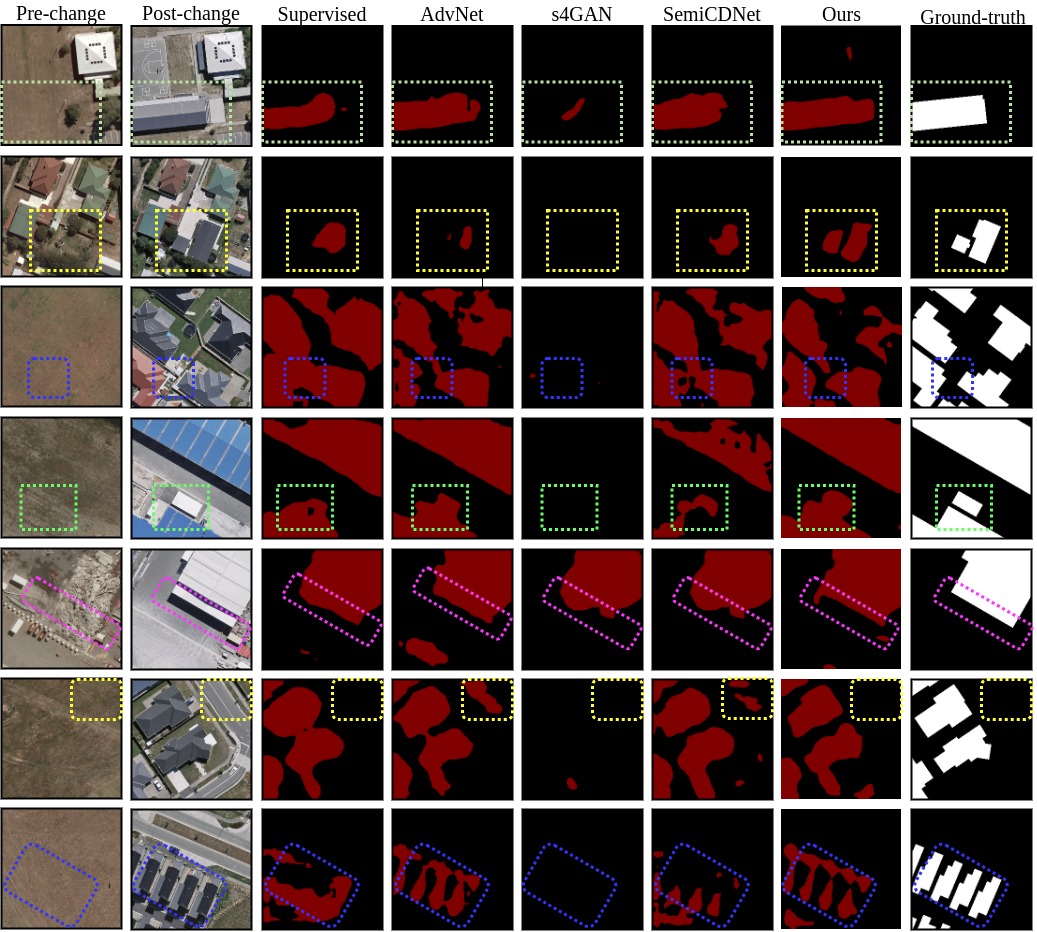}
    \caption{Additional qualitative results for WHU(5\%)$\rightarrow$WHU.}
    \label{fig:whu5-whu}
\end{figure}

Figure \ref{fig:whu5-whu} shows additional qualitative examples for WHU(5\%)$\rightarrow$WHU. From these examples also we can see that proposed semi-supervised approach generates much better change maps than the SOTA methods, further demonstrating its effectiveness of leveraging information from freely-available, unlabeled, remote-sensing data to improve the CD performance.

\clearpage
\subsection{Additional qualitative examples for WHU(10\%)$\rightarrow$WHU}
\begin{figure}[!h]
    \centering
    \includegraphics[width=\linewidth]{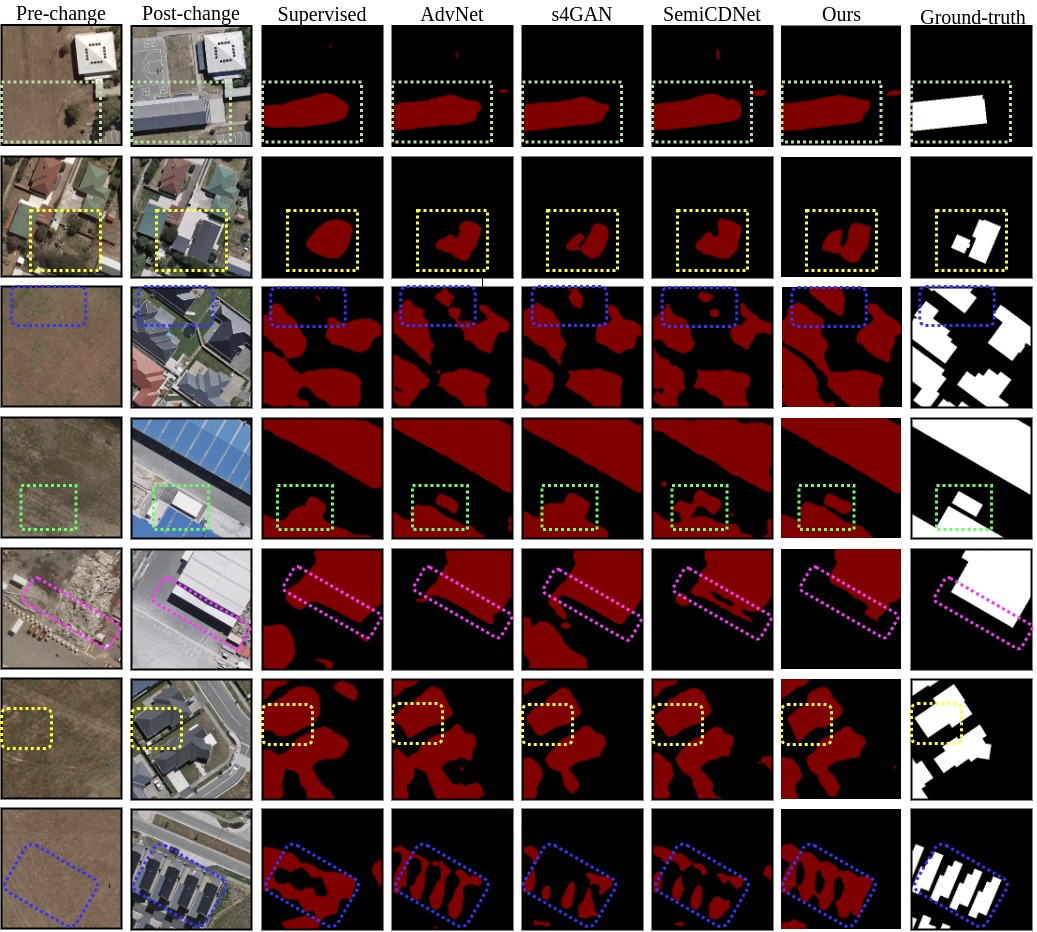}
    \caption{Additional qualitative results for WHU(10\%)$\rightarrow$WHU.}
    \label{fig:whu10-whu}
\end{figure}
Figure \ref{fig:whu10-whu} shows additional qualitative examples for WHU(5\%)$\rightarrow$WHU. Similar to Fig. \ref{fig:whu5-whu}, from these examples as well, we can see that the proposed semi-supervised approach generates much better change maps than SOTA methods, further demonstrating its effectiveness in leveraging freely-available, unlabeled remote-sensing images to improve CD performance.

\clearpage
\subsection{Additional qualitative examples for WHU(20\%)$\rightarrow$WHU}
\begin{figure}[!h]
    \centering
    \includegraphics[width=\linewidth]{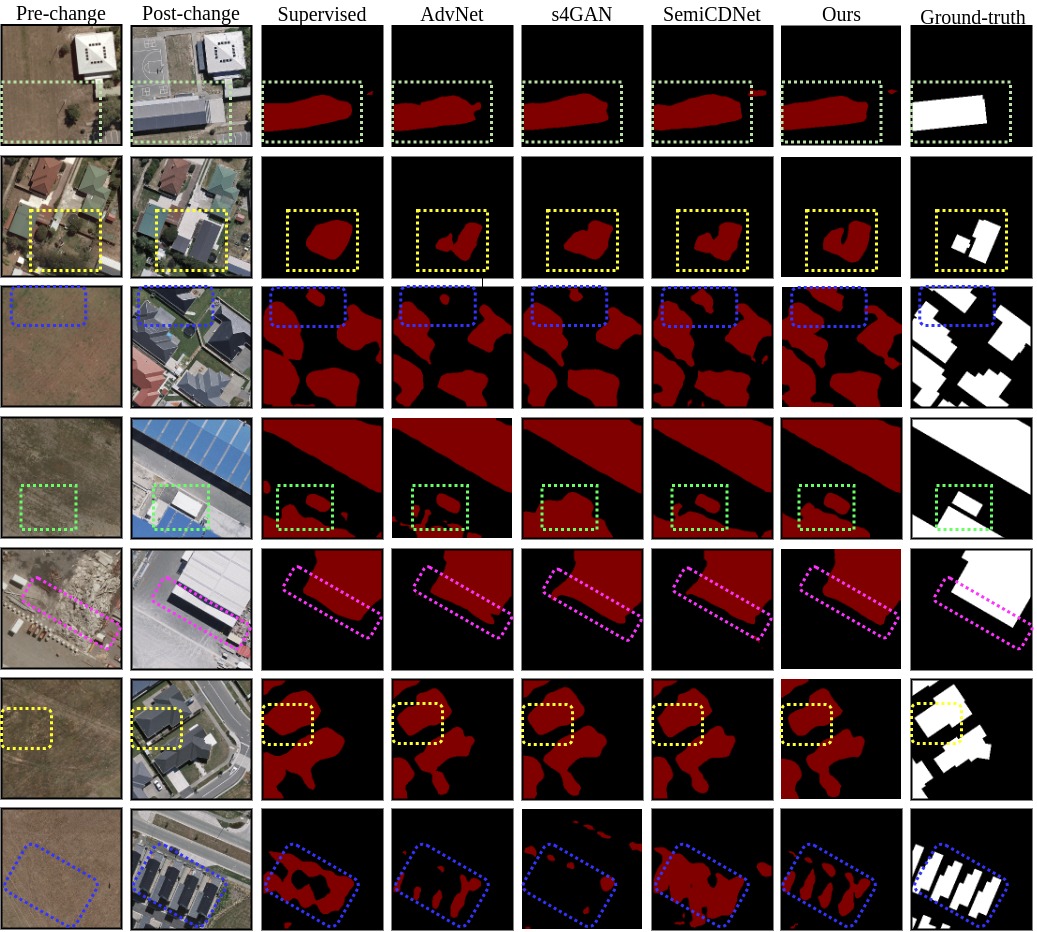}
    \caption{Additional qualitative results for WHU(20\%)$\rightarrow$WHU.}
    \label{fig:levir20-levir}
\end{figure}
Figure \ref{fig:levir20-levir} shows additional qualitative examples for WHU(20\%)$\rightarrow$WHU. Similar to Fig. \ref{fig:levir5-levir} and \ref{fig:levir10-levir},  from these examples as well, we can see that the proposed semi-supervised approach generates much better change maps than SOTA methods, further demonstrating its effectiveness in leveraging freely-available, unlabeled remote-sensing images to improve CD performance.

\clearpage
\subsection{Additional qualitative examples for WHU(40\%)$\rightarrow$WHU}
\begin{figure}[!h]
    \centering
    \includegraphics[width=\linewidth]{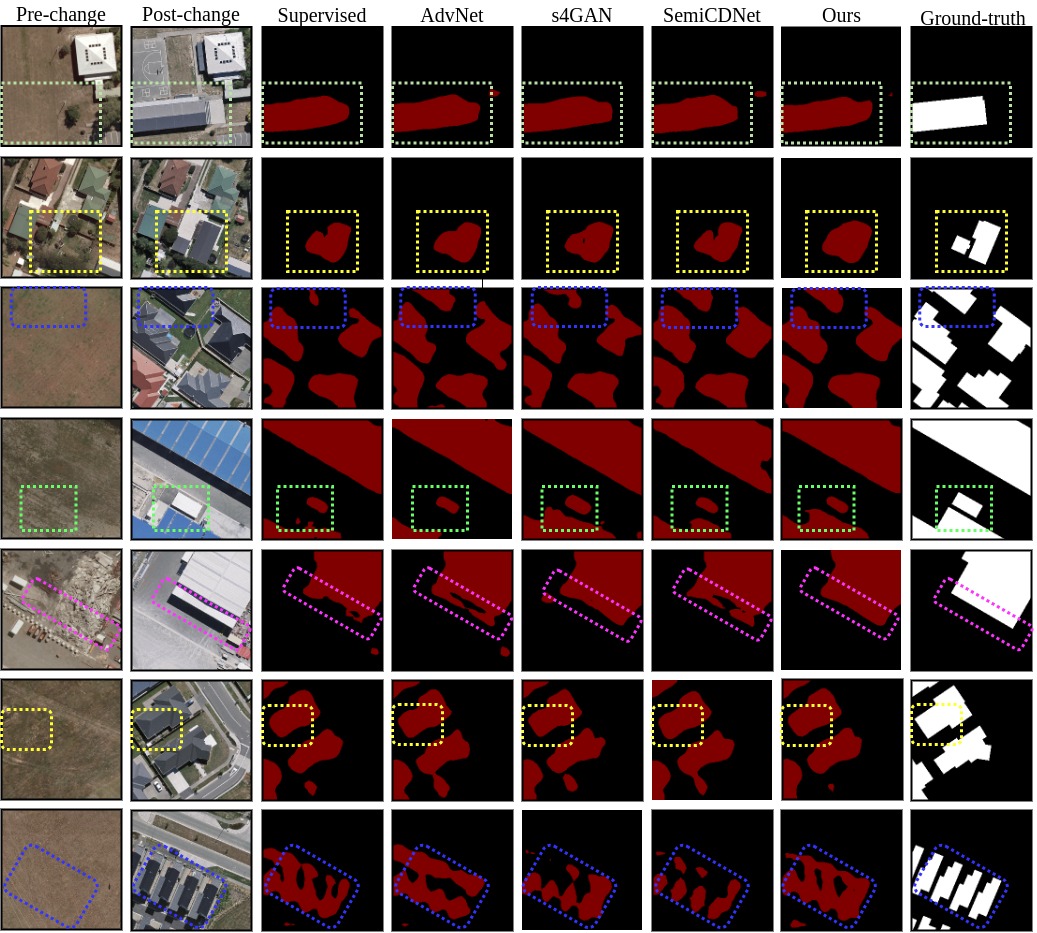}
    \caption{Additional qualitative results for WHU(40\%)$\rightarrow$WHU.}
    \label{fig:levir40-levir}
\end{figure}
Figure \ref{fig:levir40-levir} shows additional qualitative examples for WHU(40\%)$\rightarrow$WHU. Similar to Fig. \ref{fig:levir5-levir}, \ref{fig:levir10-levir}, and \ref{fig:levir20-levir},  from these examples as well, we can see that the proposed semi-supervised approach generates much better change maps than SOTA methods, further demonstrating its effectiveness in leveraging freely-available, unlabeled remote-sensing images to improve CD performance.

\clearpage
\section{Additional qualitative examples for model transferability: LEVIR$\rightarrow$WHU}
In this section, we provide additional qualitative results related to model transferability experiments. More specifically, we show the qualitative results when model is trained on LEVIR and test on WHU for different percentages of labeled data used in the training.
\subsection{Examples for LEVIR(5\%)$\rightarrow$WHU}
\begin{figure}[!h]
    \centering
    \includegraphics[width=\linewidth]{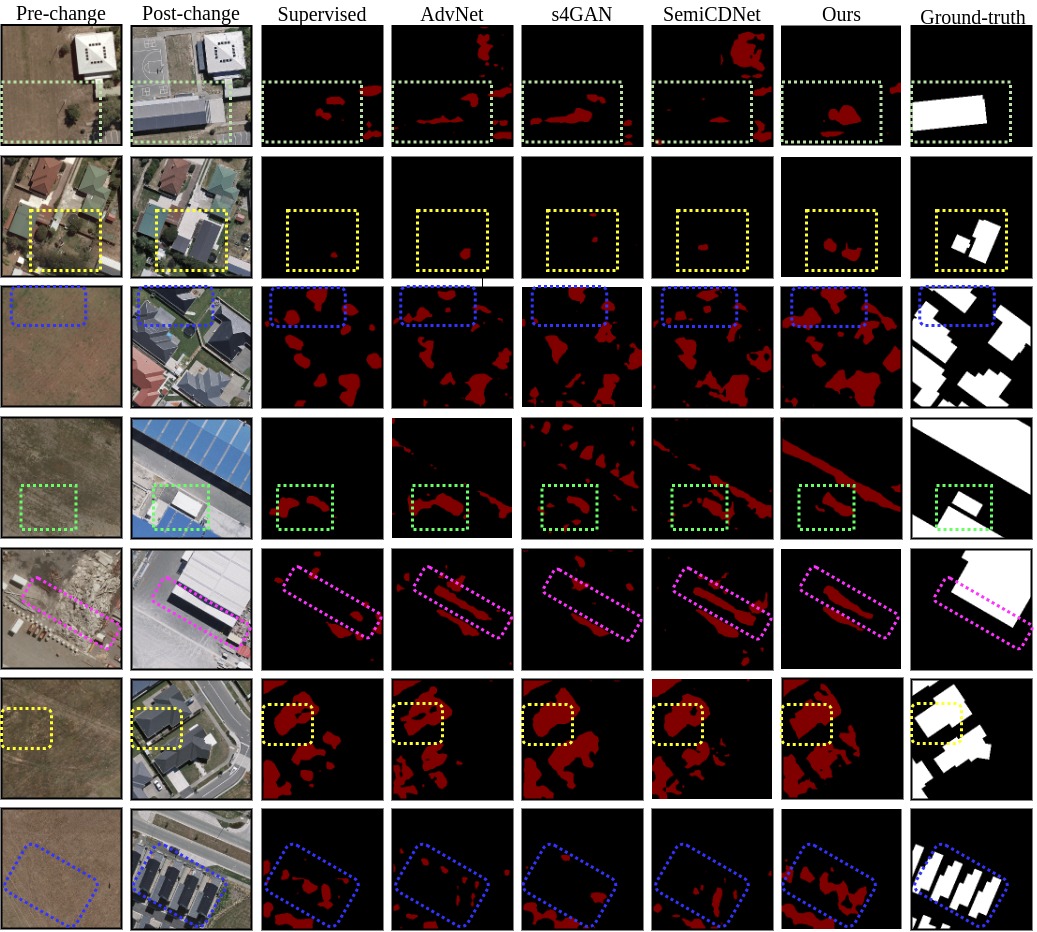}
    \caption{Additional qualitative results for model transferability: LEVIR(5\%)$\rightarrow$WHU.}
    \label{fig:levir5-whu}
\end{figure}

\clearpage
\subsection{Examples for LEVIR(10\%)$\rightarrow$WHU}
\begin{figure}[!h]
    \centering
    \includegraphics[width=\linewidth]{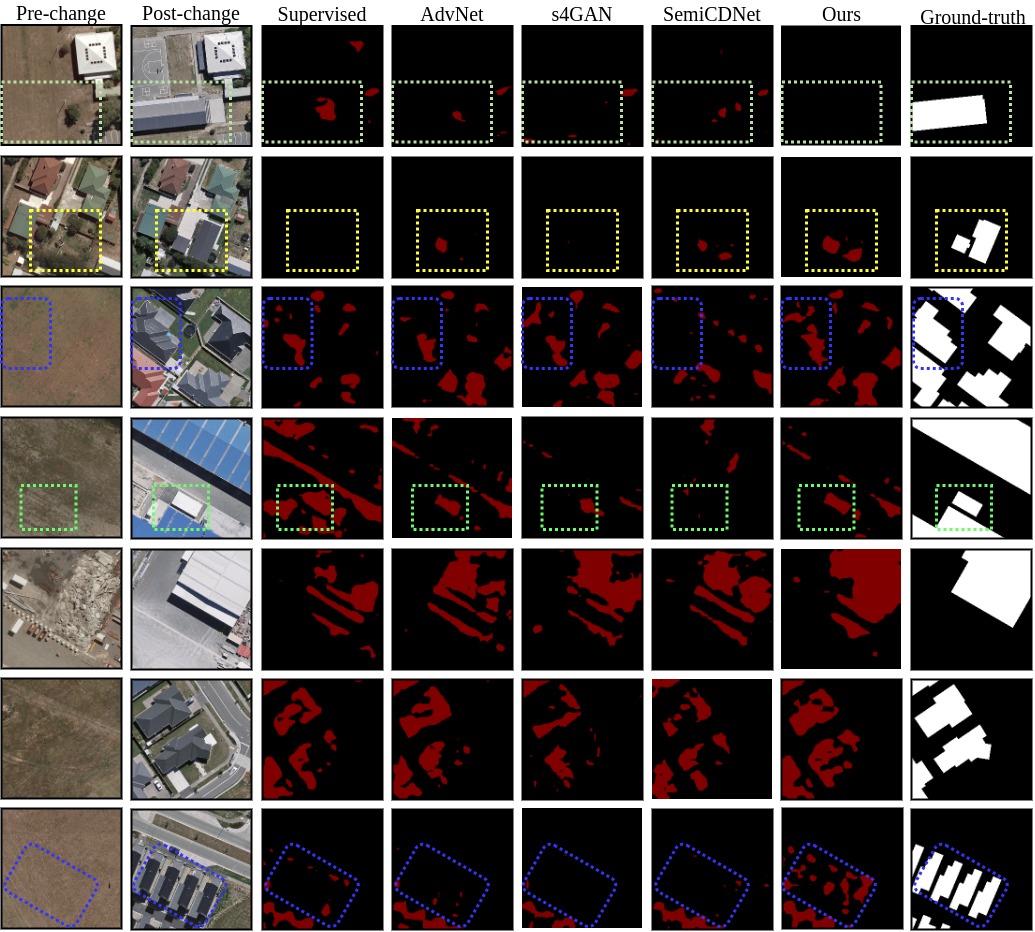}
    \caption{Additional qualitative results for model transferability: LEVIR(10\%)$\rightarrow$WHU.}
    \label{fig:levir10-whu}
\end{figure}

\clearpage
\subsection{Examples for LEVIR(20\%)$\rightarrow$WHU}
\begin{figure}[!h]
    \centering
    \includegraphics[width=\linewidth]{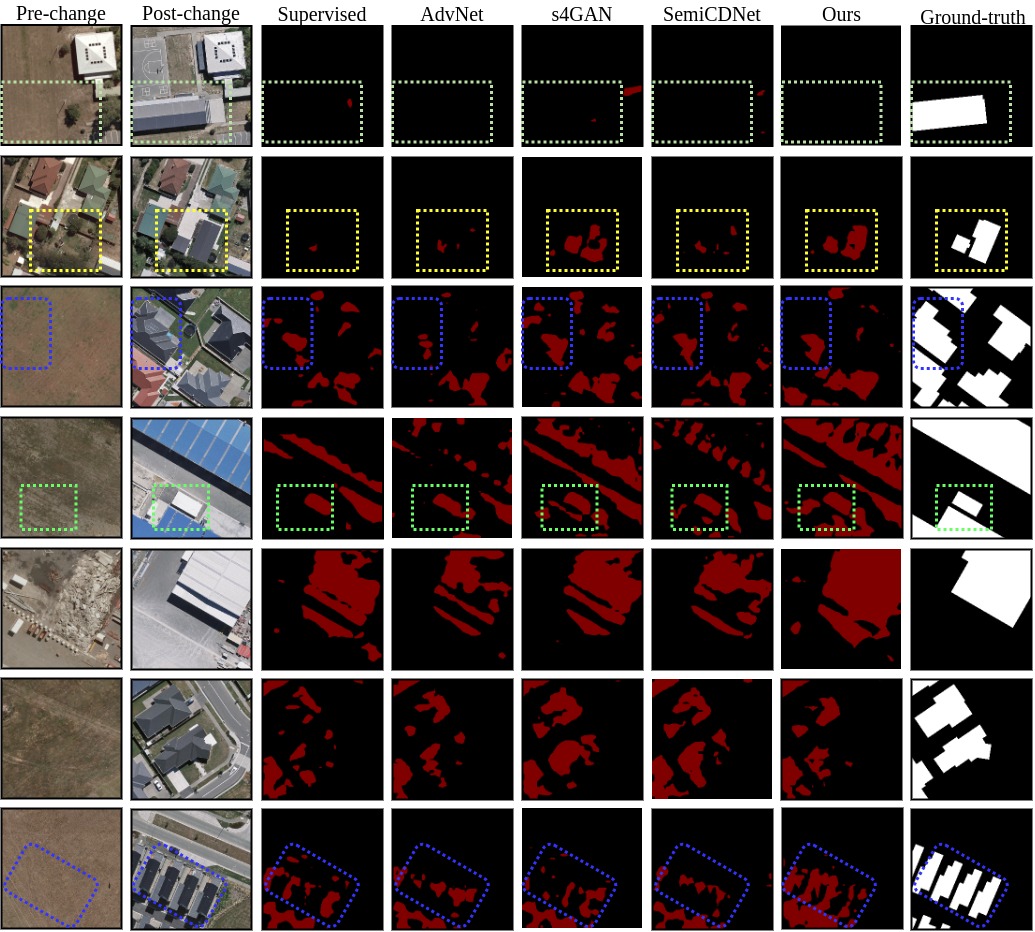}
    \caption{Additional qualitative results for model transferability: LEVIR(20\%)$\rightarrow$WHU.}
    \label{fig:levir20-whu}
\end{figure}

\clearpage
\subsection{Examples for LEVIR(40\%)$\rightarrow$WHU}
\begin{figure}[!h]
    \centering
    \includegraphics[width=\linewidth]{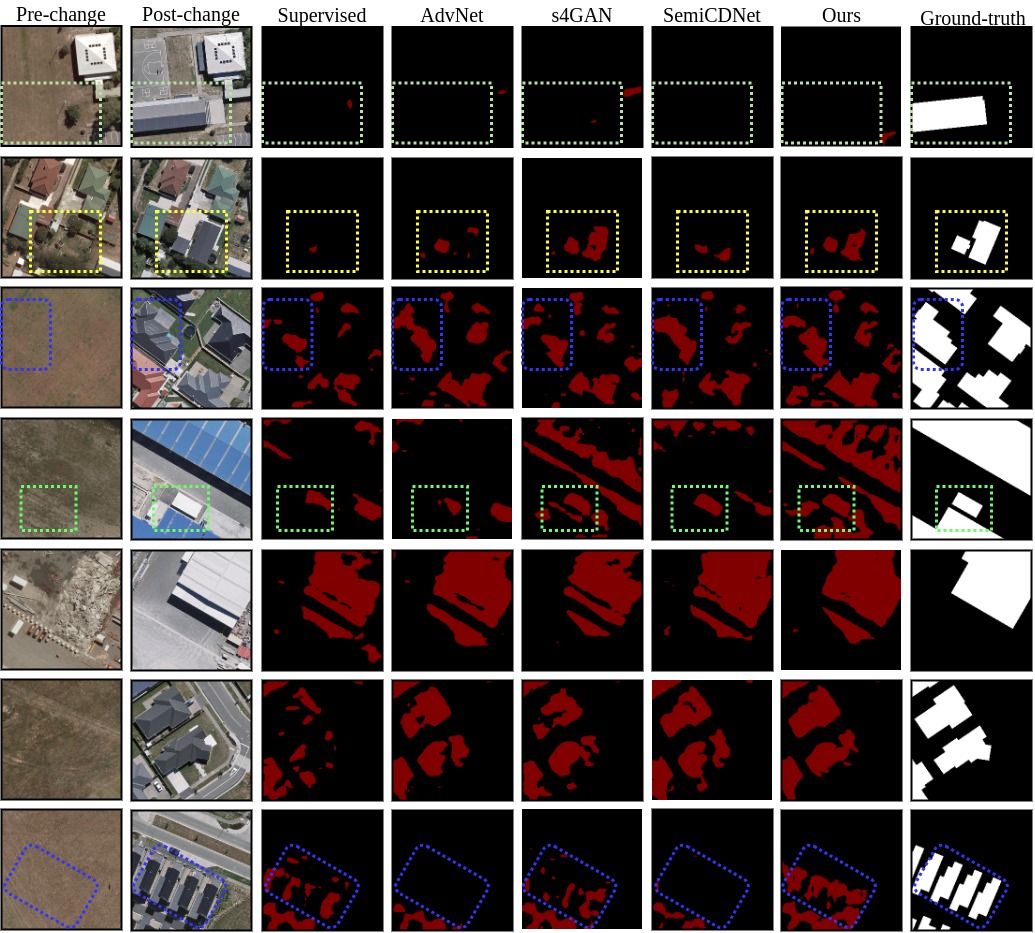}
    \caption{Additional qualitative results for model transferability: LEVIR(40\%)$\rightarrow$WHU.}
    \label{fig:levir40-whu}
\end{figure}

\clearpage
\section{Additional qualitative results for cross-domain experiments}
In this section, we present the additional qualitative results for cross-domain experiments. We combine labeled and unlabeled data from two different datasets and evaluate the performance. More specifically, we use different percentage of labeled data from LEVIR and consider unlabeled data from WHU into the semi-supervised learning process.

\subsection{Additional qualitative examples for \{LEVIR(5\%),WHU(unsup.)$\rightarrow$LEVIR\}}
\begin{figure}[!h]
    \centering
    \includegraphics[width=\linewidth]{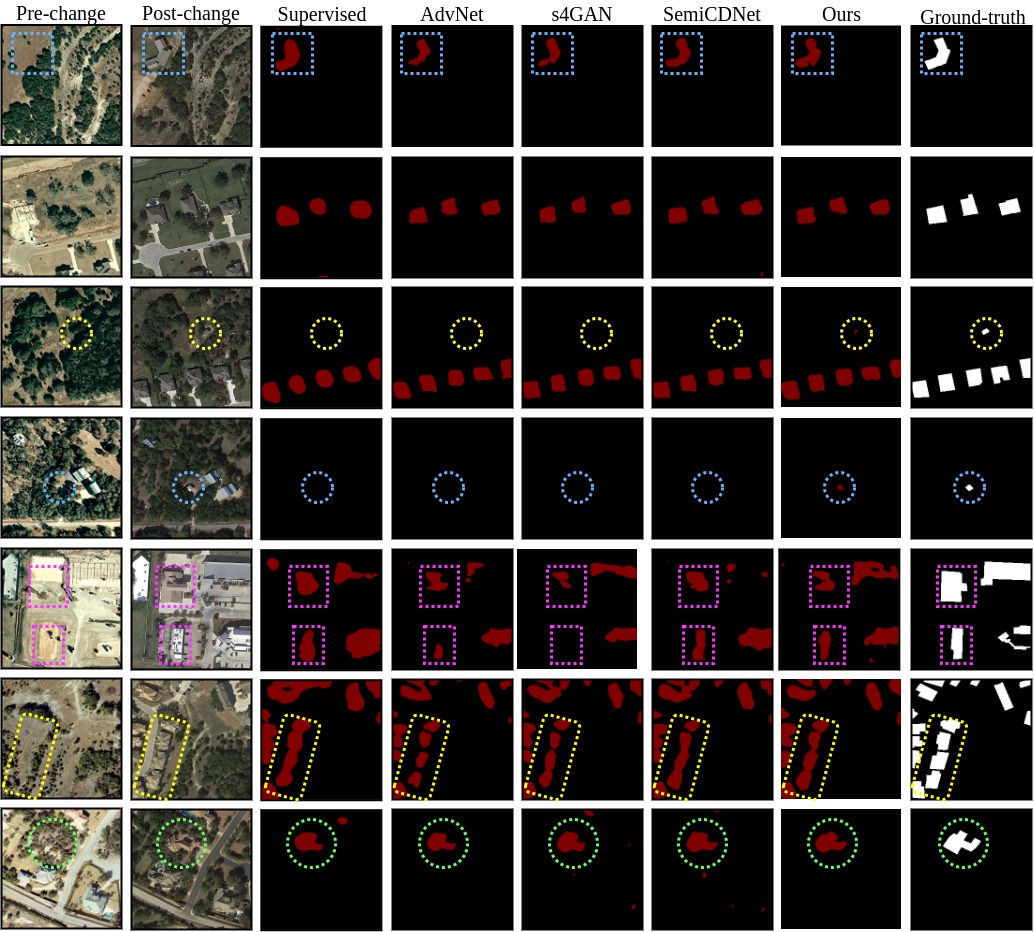}
    \caption{Additional qualitative results for model transferability: LEVIR(5\%)$\rightarrow$WHU.}
    \label{fig:levir40-whu}
\end{figure}

\clearpage
\subsection{Additional qualitative examples for \{LEVIR(10\%),WHU(unsup.)$\rightarrow$LEVIR\}}
\begin{figure}[!h]
    \centering
    \includegraphics[width=\linewidth]{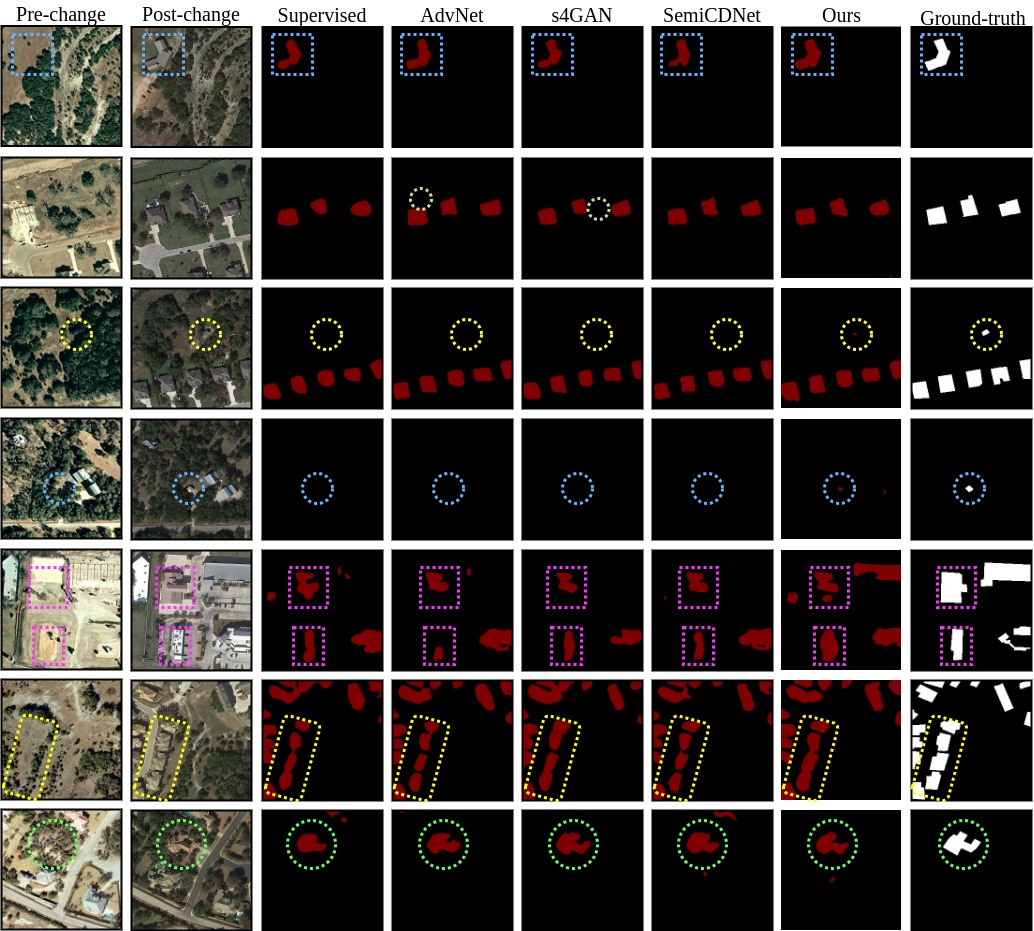}
    \caption{Additional qualitative results for model transferability: LEVIR(10\%)$\rightarrow$WHU.}
    \label{fig:levir40-whu}
\end{figure}

\clearpage
\subsection{Additional qualitative examples for \{LEVIR(20\%),WHU(unsup.)$\rightarrow$LEVIR\}}
\begin{figure}[!h]
    \centering
    \includegraphics[width=\linewidth]{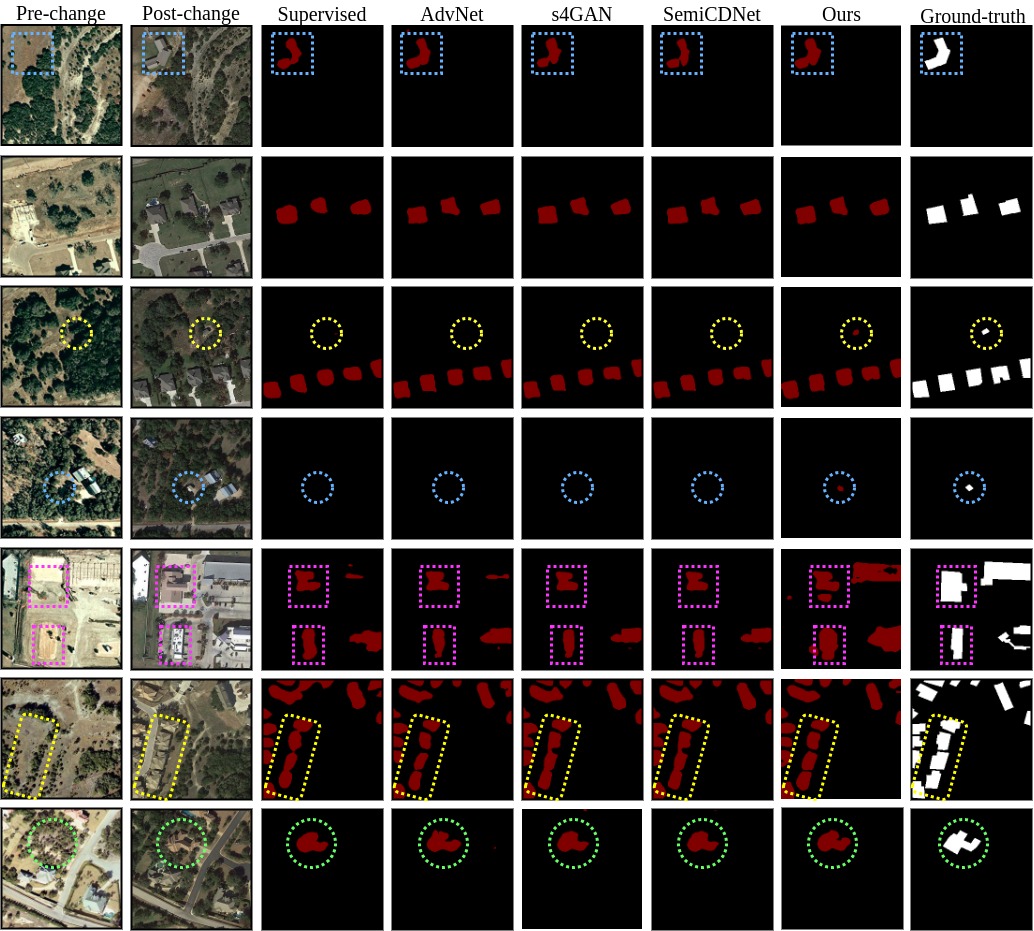}
    \caption{Additional qualitative results for model transferability: LEVIR(20\%)$\rightarrow$WHU.}
    \label{fig:levir40-whu}
\end{figure}

\clearpage
\subsection{Additional qualitative examples for \{LEVIR(40\%),WHU(unsup.)$\rightarrow$LEVIR\}}
\begin{figure}[!h]
    \centering
    \includegraphics[width=\linewidth]{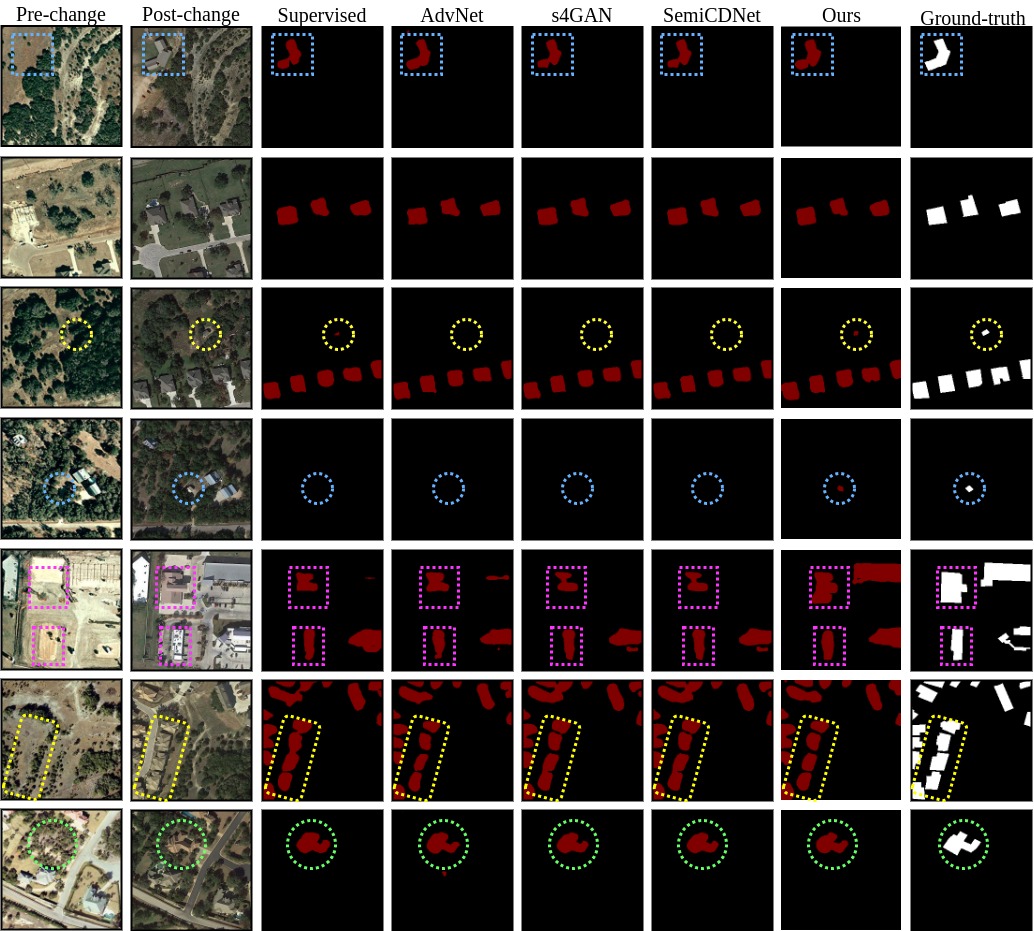}
    \caption{Additional qualitative results for model transferability: LEVIR(40\%)$\rightarrow$WHU.}
    \label{fig:levir40-whu}
\end{figure}
